\documentclass[preprint,authoryear, english,fleqn]{elsarticle}
\usepackage{enumerate}
\usepackage{amsmath}
\usepackage{multirow}
\usepackage{threeparttable}
\usepackage{diagbox}
\usepackage[T1]{fontenc}
\usepackage{array}
\usepackage{longtable}
\usepackage{graphicx}
\usepackage{esint}
\usepackage{float}%%
\usepackage{caption}
\usepackage{subcaption}
\usepackage{amssymb}%nrightarrow
\usepackage{amsthm}
\usepackage{bbm}

\usepackage{array,booktabs}
\usepackage{verbatim}

\usepackage{color}
\makeatletter
\AtBeginDocument{}
\providecommand{\\}{\\}
\@ifundefined{date}{}{\date{}}
\usepackage{epstopdf}
\DeclareGraphicsExtensions{.pdf,.jpeg,.png}
\usepackage[paperheight=11.69in,paperwidth=8.27in,left=1.2in,right=1.2in,top=1.2in,bottom=1.2in,headheight=1.2in]{geometry}
\usepackage[hidelinks]{hyperref}

\def\equationautorefname~#1\null{Eq.~(#1)\null}
\usepackage{cleveref}
\crefrangeformat{equation}{Eqs. #3(#1)#4--#5(#2)#6}
\creflabelformat{equation}{#2\textup{#1}#3}
\usepackage{longtable}
\makeatother
\usepackage{babel}
\usepackage{graphicx}
\graphicspath{ {./figure/} }
\allowdisplaybreaks[4]
\usepackage[textfont=normalsize]{subcaption}
\usepackage[font=normalsize]{caption}
\usepackage{enumitem}
\setlist[enumerate,1]{label=\textup{(\roman{*})}, parsep = 0pt, itemindent=0.1cm, itemsep=0pt}
\usepackage{adjustbox}
\usepackage{tabularx}
\usepackage{tabu}
\usepackage{color}

\makeatletter% delete the submit to elsvier
\def\ps@pprintTitle{%
   \let\@oddhead\@empty
   \let\@evenhead\@empty
   \def\@oddfoot{\reset@font\hfil\thepage\hfil}
   \let\@evenfoot\@oddfoot
}
\makeatother
\def\eqref#1{{(\ref{#1})}}

\usepackage[mathlines,running]{lineno}
\expandafter\let\expandafter\oldequation\csname equation\endcsname
\expandafter\let\expandafter\oldgather\csname gather\endcsname
\expandafter\let\expandafter\endoldequationstar\csname endequation*\endcsname
\expandafter\def\csname equation\endcsname{%
  \ifLineNumbers%
  \expandafter\linenomath%
  \fi%
  \oldequation%
}
\expandafter\def\csname gather\endcsname{%
  \ifLineNumbers%
  \expandafter\linenomath%
  \fi%
  \oldgather%
}
\expandafter\def\csname endequation*\endcsname{%
  \endoldequationstar%
  \ifLineNumbers%
  \def\maybeendlinenomath{\expandafter\endlinenomath}%
  \else
  \def\maybeendlinenomath{}%
  \fi%
  \expandafter\maybeendlinenomath%
}

\begin{document}
\begin{frontmatter}{}
\title{Unsupervised Knowledge Adaptation for Passenger Demand Forecasting}
% \let\WriteBookmarks\relax
% \def\floatpagepagefraction{1}
% \def\textpagefraction{.001}
%\shorttitle{}
%\shortauthors{Li et al.}

%\begin{frontmatter}

\author[a]{Can Li}
\address[a]{School of Computer Science and Engineering, University of New South Wales, Sydney, NSW 2052, Australia}
\author[b]{Lei Bai}
\address[b]{School of Electrical and Information Engineering, University of Sydney, Sydney, NSW 2008, Australia}
\author[c]{Wei Liu\corref{correspond_author}}\cortext[correspond_author]{Corresponding author}
\ead{wei.w.liu@polyu.edu.hk}
\address[c]{Department of Aeronautical and Aviation Engineering, The Hong Kong Polytechnic University, Hong Kong, China}
\author[a]{Lina Yao}
\author[d]{S Travis Waller}
\address[d]{Research Centre for Integrated Transport Innovation, School of Civil and Environmental Engineering, University of New South Wales, Sydney, NSW 2052, Australia}

\begin{abstract}
Considering the multimodal nature of transport systems (e.g., train, bus) and potential cross-modal correlations (e.g., similar temporal patterns), there is a growing trend of enhancing demand forecasting accuracy by learning from multimodal data, i.e., forecasting models that are optimized with data from multiple transport modes jointly. These multimodal forecasting models are able to improve accuracy, but they can be less practical when different parts of multimodal datasets are owned by different institutions who cannot directly share data among them. While various institutions may not be able to share their data with each other directly, they may share forecasting models trained by their data, where such models cannot be used to identify the exact information from their datasets. In this context, this study proposes an \textbf{Un}supervised \textbf{K}nowledge \textbf{A}daptation \textbf{D}emand \textbf{F}orecasting  (\textbf{Un-Kadf}) framework, which forecasts the demand of one mode (i.e., the target mode) by utilizing a pre-trained model based on data of another mode, but does not require direct data sharing of another transport mode (i.e., the source mode). The proposed framework utilizes the potential shared patterns among multiple transport modes for improving forecasting performance while avoiding the direct sharing of data among different institutions. Specifically, a pre-trained forecasting model is first learned based on the data of a source mode, which can capture and memorize the source travel patterns (but not the exact source data). Then, the demand data of the target dataset is encoded into an individual knowledge part and a sharing knowledge part which will extract travel patterns by two networks, i.e., individual extraction network and sharing extraction network, respectively. The unsupervised knowledge adaptation strategy is utilized to form the sharing features for further forecasting by making the pre-trained network and the sharing extraction network analogous.
Extensive experiments conducted on real-world datasets from the Greater Sydney area covering four public transit modes (i.e., bus, train, light rail, and ferry) demonstrate that the proposed approach outperforms a number of baseline methods and state-of-the-art models. 
Our findings also illustrate that unsupervised knowledge adaptation by sharing the pre-trained model to the target transport mode can improve the forecasting performance without the dependence on direct data sharing. 
\end{abstract}


\end{frontmatter}{}

\section{Introduction} \label{sec: introduction}
Artificial intelligence and machine learning algorithms have now been widely adopted for demand forecasting given the increasing availability of multi-source datasets \citep{chen2020nonconvex}. Earlier studies often utilize statistical time-series methods and traditional machine learning methods to explore the temporal information for traffic prediction, such as Auto-Regressive Integrated Moving Average (ARIMA) \citep{lippi2013short}, Kalman Filter \citep{xue2015short}, and Support Vector Machine (SVM) \citep{feng2018adaptive}. Given its strong capability to deal with non-linear relations and large-scale data, deep learning has attracted substantial attention for demand prediction in recent years. Specifically, Recurrent Neural Network (RNN) and its variants such as Long Short-Term Memory (LSTM) \citep{kim2020stepwise} and Gated Recurrent Unit (GRU) \citep{peng2021cnn} are used for temporal relations extraction and analysis. To further explore the spatial correlations, Convolution Neural Network (CNN) \citep{ma2018parallel, liu2019contextualized, zhang2021short} and Graph Convolution Network (GCN) \citep{zhang2019multistep, cui2020learning, guo2020optimized,  tang2021multi} are utilized to characterize spatial features, which help improve the demand/traffic prediction accuracy.

\begin{figure}[htbp]
\centering
\begin{minipage}[t]{0.49\textwidth}
\centering
\includegraphics[width=6cm]{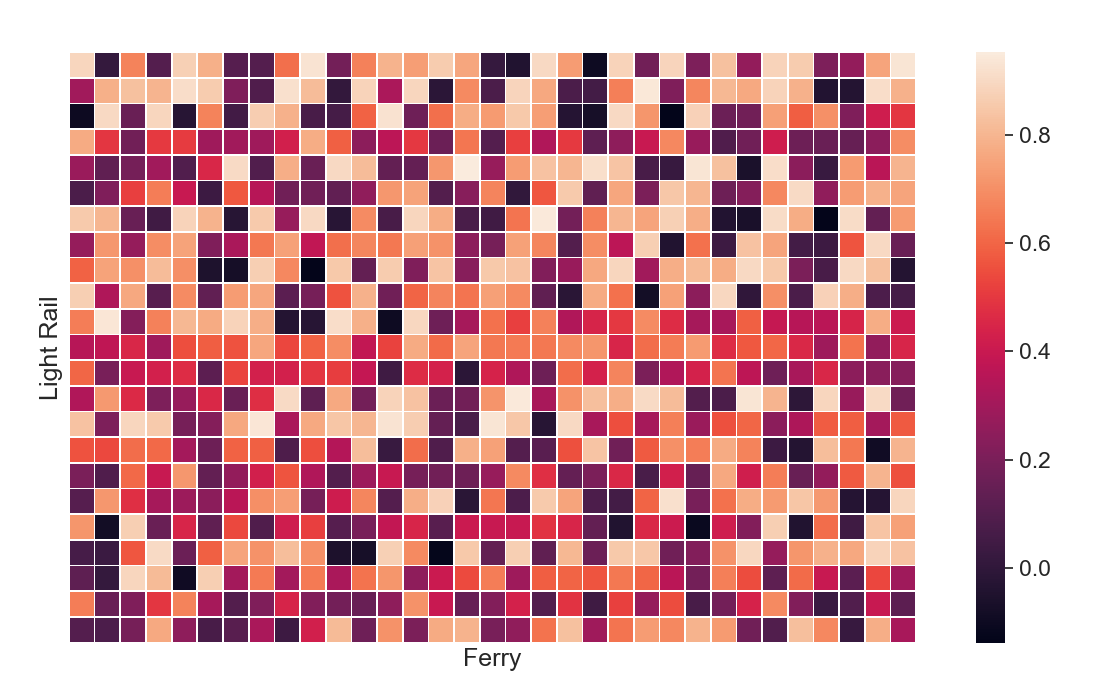}
\caption{Correlation Coefficient Heatmap of the Light Rail and Ferry Demands in Sydney}
\label{fig:heatmap}
\end{minipage}
\begin{minipage}[t]{0.49\textwidth}
\centering
\includegraphics[width=6cm]{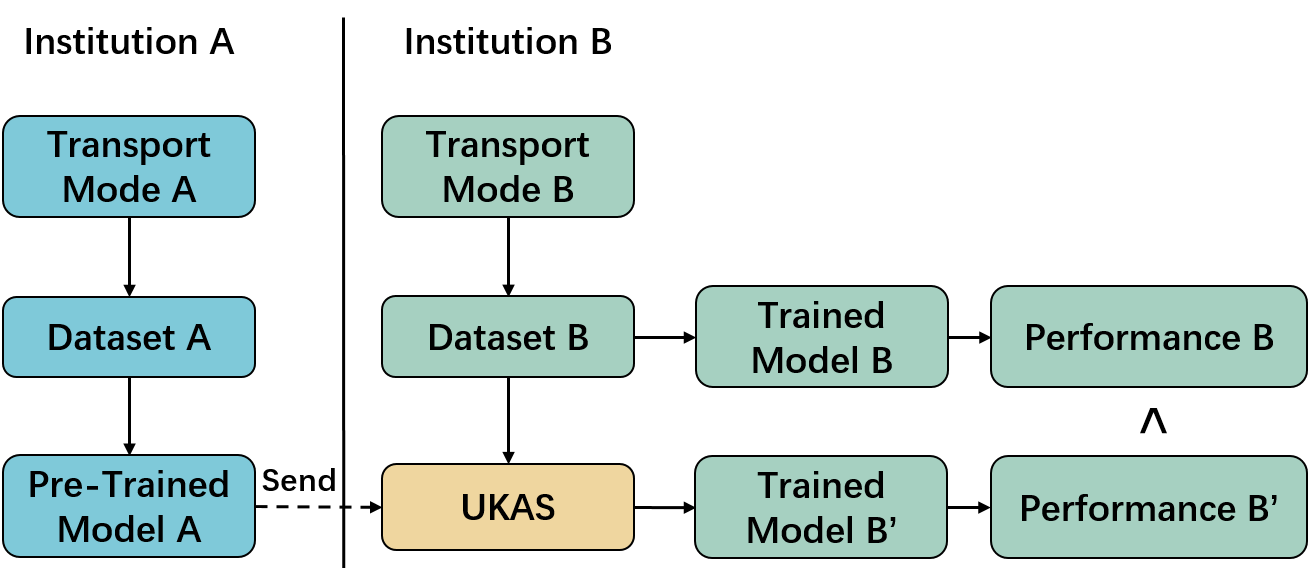}
\caption{The Framework of the Unsupervised Knowledge Adaptation Strategy (UKAS) via Model Sharing}
\label{fig:flowchart}
\end{minipage}
\end{figure}

While the aforementioned works have achieved competent prediction accuracy, they only focus on the exploration of a single transport mode and do not utilize the potential similarities/correlations among different transport modes for better prediction. Modern transport systems are in nature multimodal, including various modes such as the bus, train, and light rail. There are potential similar or correlated travel patterns among different modes. For instance, Fig.~\ref{fig:heatmap} presents the Pearson correlation coefficient for the demand at any ferry station and the demand at any light rail station in Sydney, where a lighter color means a more significant positive correlation. As can be observed, nearly 50\% pairs of ferry stations and light rail stations have a relatively high Pearson correlation coefficient (i.e., larger than $0.8$), indicating the strong correlations between these two public transit modes.\footnote{The detailed correlation information of the evaluated four transport modes collected from the Greater Sydney area (i.e., bus, train, light rail, and ferry) will be explained in Subsection~\ref{sec:dataset}.} Recognizing the correlations/similarities among various transport modes, some studies \citep{ye2019co, toman2020dynamic, li2021multi, ke2021joint} try to co-predict demand of multiple modes by utilizing the multimodal datasets. The experimental results of these studies illustrate that the knowledge adaptation via data sharing among various transport modes is able to improve the forecasting accuracy. However, in practice, different parts (e.g., data of different modes operated by different public/private operators) of the multimodal dataset might be owned by different institutions. There might be constraints on directly sharing data among these different institutions. For example, in Hong Kong, five different companies provide bus services\footnote{https://en.wikipedia.org/wiki/Bus\_services\_in\_Hong\_Kong}, and Mass Transit Railway (MTR) is in charge of the subway operation\footnote{https://www.mtr.com.hk/en/customer/main/}. These institutions may not be able to directly share their detailed data with each other due to either privacy concerns or other practical constraints. This motivates the current study to explore new directions for enhancing passenger demand prediction that only requires indirect/limited input from other sources and does not require direct access to details of other datasets. 

In particular, this study aims to improve the demand forecasting performance of the target transport mode with the mechanism of unsupervised knowledge adaptation by model sharing instead of direct data sharing in the context of the multimodal transport system in a city. To make full use of the (indirect/limited) information from the source transport mode to enhance the forecasting performance of the target transport mode, this study proposes an \textbf{Un}supervised \textbf{K}nowledge \textbf{A}daptation \textbf{D}emand \textbf{F}orecasting  (\textbf{Un-Kadf}) framework for station-level (or location-based) demand forecasting. To illustrate the mechanism of model sharing between two transit modes, Fig.~\ref{fig:flowchart} displays the framework of the model sharing. Given the potential correlations among two datasets collected from two transport modes operated by different institutions, the two datasets can be used as the source dataset for each other, i.e., Dataset A is regarded as the source dataset for target Dataset B and vice visa. First, a pre-trained demand forecasting Model A is optimized by Dataset A. The mechanism of such a pre-trained model has been used in natural language process (NLP) \citep{devlin2019bert} and computer vision (CV) \citep{tan2019efficientnet}, which can help dramatically improve the model performance. Then, Model A can be employed in the optimization process of Model B on Dataset B via a model sharing strategy to get Model B$'$, which could improve the prediction accuracy (i.e.,  Performance B$'$ is better than Performance B where Performance B is obtained only based on its own data). The learning process of Model B$'$ does not need direct access to any data from Dataset A. Instead, the model sharing strategy adapts the knowledge from A to B in an unsupervised manner (i.e., unsupervised knowledge adaption) \citep{pan2009survey}.

To achieve performance improvement by unsupervised knowledge adaptation via model sharing, several challenges have to be addressed. First, different public transit modes have different numbers of stations so that the temporal information of them cannot be directly extracted by recurrent networks with the same structure. An adaptable recurrent network has to be developed that can handle such differences between the source and target datasets without destroying travel patterns. Second, unnecessary task-specific features from the source dataset may influence the forecasting results of the target dataset, which should be avoided. In particular, this study designs an encoder-decoder structure with LSTM to extract valuable knowledge from the source dataset and ensure that the adaptable pre-trained LSTM structure can be applied to the target dataset. Based on the encoder-decoder framework, the target dataset is encoded into an individual knowledge part and a sharing knowledge part. The individual part is analyzed by the individual extraction recurrent network, while the sharing part is explored by the sharing extraction recurrent network. Then, we adopt the transfer learning strategy for unsupervised knowledge adaptation to optimize the sharing extraction network based on the sharing knowledge part of the target dataset, which can utilize the information of the source dataset memorized in the adaptable LSTM structure. The aim of transfer learning is to develop a new model for the target distribution by transferring relative knowledge based on the source distribution \citep{pan2009survey}.

The main contributions of this paper are summarized in the following. 
(i) This study proposes a new approach to enhance the passenger demand forecasting performance for multimodal transport systems by unsupervised knowledge adaptation from the source dataset to the target dataset without the dependence of direct data sharing.
(ii) The proposed \textbf{Un-Kadf} framework helps address the unsupervised knowledge adaptation problem via designing the pre-trained recurrent network optimized by the source dataset and sharing the pre-trained network to the target dataset for further optimization.
(iii) This study conducts comprehensive experiments on large-scale real-world public transport datasets with four modes. The results show that the proposed model significantly outperforms existing methods and demonstrate the potential of boosting multimodal demand forecasting accuracy by unsupervised knowledge adaptation via model sharing.

The rest of this paper is organized as follows. We introduce the related literature in Section~\ref{sec:related work}. We then discuss the case study dataset and define the demand prediction problems in Section~\ref{sec:datasetanalysis}. Section~\ref{sec:model} presents the proposed \textbf{Un-Kadf} model. The evaluation of the proposed method and comparison with other existing methods are presented in Section~\ref{sec:experiment}. Finally, Section~\ref{sec:conclusion} concludes this paper.

\section{Related Work} \label{sec:related work}

This section first reviews relevant data-driven studies on demand forecasting (both single-mode and multi-mode). Then, transport forecasting problems dealing with knowledge adaptation methods are discussed.

\subsection{Demand Forecasting under Single Travel Mode}

Given the increasing availability of transport datasets, data-driven demand forecasting has been received much attention in recent years \citep{ma2018estimating}. A large number of studies examine the estimation of demand with the data collected from the one transit mode in concern via the usage of statistical time-series methods or machine learning strategies, especially deep-learning-based models.

In particular, at earlier stages, studies for demand prediction are mainly based on statistical methods and traditional time-series regression models. For instance, Auto-Regressive Integrated Moving Average (ARIMA) \citep{lippi2013short}, Kalman Filter \citep{xue2015short}, Support Vector Machine (SVM) \citep{feng2018adaptive}, Vector Auto-regression \citep{cheng2022real} and their variants \citep{moreira2013predicting} have been widely utilized. However, these strategies have limited capability to deal with non-linear temporal correlations and large-scale datasets for precise demand forecasting.

To extract non-linear temporal information for large-scale datasets, deep-learning-based models are adopted for prediction problems under single target transport mode, including Fully Connected Layer (FCL) \citep{lv2014traffic} and RNN-based networks (e.g., LSTM and GRU) \citep{xu2017real, kim2020stepwise}. In detail, a series of AutoEncoder is stacked in \citet{lv2014traffic} to learn generic traffic flow features where AutoEncoder consists of multiple FCLs. FCLs can hardly capture the long-term knowledge, which motivates the utilization of RNN-based models for demand forecasting. For instance, both \citet{xu2017real} and \citet{kim2020stepwise} designed the predictive framework based on LSTM for taxi demand prediction.

Considering the relevance of spatial information for demand estimation, CNN is introduced for transport problems and has been combined with RNN-based models to obtain a more comprehensive understanding of spatial-temporal correlations \citep{ke2018hexagon, liu2019contextualized, guo2019deep}. Given that CNN can only handle relations among adjacent areas, GCN is further applied for non-Euclidean spatial information extraction. Similarly, a series of studies \citep{li2019learning, geng2019spatiotemporal, jin2020urban,bai2020adaptive} combine GCN and recurrent networks to capture spatial-temporal knowledge for better demand forecasting. Moreover, in order to reduce the error accumulation caused by the iteration of RNN, \citet{bai2019stg2seq} and \citet{li2020graph} stack a set of gated graph convolution layers to model spatial and temporal information simultaneously for further single-mode demand prediction. In addition, to avoid over simplistic integration of heterogeneous data, \citet{zhou2021urban} takes advantage of neural ordinary differential equations (ODE) to capture the continuous-time dynamics of the latent states for further forecasting.

\subsection{Demand Forecasting under Multiple Travel Modes} 

Correlations and similarities among multiple different transport modes provide new opportunities and dimensions to enhance the demand forecasting performance, which has been receiving growing attention recently.

\citet{ye2019co} designs a co-prediction model to predict the pick-up and drop-off demand for taxis and bikes with the incorporation of heterogeneous LSTM. \citet{ke2021joint} focuses on the demand of ride-hailing systems to predict solo and shared service rides jointly by constructing multi-graph convolutional networks. Furthermore, \citet{li2020knowledge} designs a recurrent network for demand prediction of the station-intensive travel mode and the station-sparse travel mode simultaneously to improve the forecasting accuracy for the station-sparse mode. A larger range of different transport modes are explored in \citet{toman2020dynamic}, including taxis, bikes, subways, and vehicles operated by transit network companies (TNCs). The vector autoregressive model with exogenous predictors is fit to predict the demand. Training multiple modes jointly for demand prediction enhancement has been verified in the aforementioned studies. However, different parts of multimodal datasets might be owned by different institutions who cannot directly share data among them, which is often the case in large cities with many different public and private operators. Existing multimodal demand prediction methods might not be applicable when such data sharing issues exist. Thus, this study aims to propose a demand prediction method that does not need direct sharing of data, but instead takes advantage of a pre-trained network optimized by the source dataset to boost the prediction performance of the target dataset.

\subsection{Transport Forecasting with Knowledge Adaptation}

Knowledge adaptation strategies have been adopted to improve the prediction performance of the target dataset for transport systems. As a popular knowledge adaptation method, transfer learning is to utilize previously developed and trained learning models to adapt the knowledge learned from the source dataset to help learn the information of the target dataset by hunting for similarities among them \citep{pan2009survey}.

Several works have explored transfer learning in the transport prediction area. However, they mainly target on the data scarcity problem caused by the unbalanced development levels of different cities and depend on the direct access to all the relevant datasets \citet{wang2019cross, yao2019learning}. In detail, \citet{wang2019cross} transfers the knowledge from a data-rich (e.g., with 90 days data) source city to a data-scarce (e.g., with 10 days data) target city by learning an inter-city region matching function to match two similar regions for crowd flow prediction. Meta-learning \citep{finn2018meta} is adopted in \citet{yao2019learning} to apply the knowledge learned from multiple cities to increase the transferring stability for spatial-temporal forecasting of the target city. Similarly, \citet{li2021domain, li2021transferability} improve the performance of the traffic flow prediction by analyzing similarities among multiple links in the highway. \citet{li2021multi} develops the memory-augmented recurrent network to adapt knowledge from the station-intensive mode to the station-sparse mode to improve the forecasting accuracy for the multimodal public transport system where the network is optimized by two datasets simultaneously. These methods have shown their capability to improve the forecasting performance for target datasets, which rely on full direct access to the source dataset and cannot be applied to dealing with the scenario with no direct access to the source dataset. Different from these works, we adopt the method of unsupervised knowledge adaptation to improve the prediction accuracy of the target dataset by adapting the knowledge memorized in the pre-trained model learned from the source dataset, which does not require direct sharing of the source dataset.

\section{Dataset and Demand Forecasting Problem Formulation} \label{sec:datasetanalysis}

This section introduces the dataset collected in the Greater Sydney area that contains multiple transit modes. Then, the possible correlations/similarities among various transport modes are illustrated, highlighting the potential of knowledge adaptation from the source mode to help enhance the forecasting performance of the target mode.

\subsection{Dataset and Mode Correlation} \label{sec:dataset}

\begin{figure}[htb]
\centering
\begin{subfigure}{.48\textwidth}
\includegraphics[width=\linewidth]{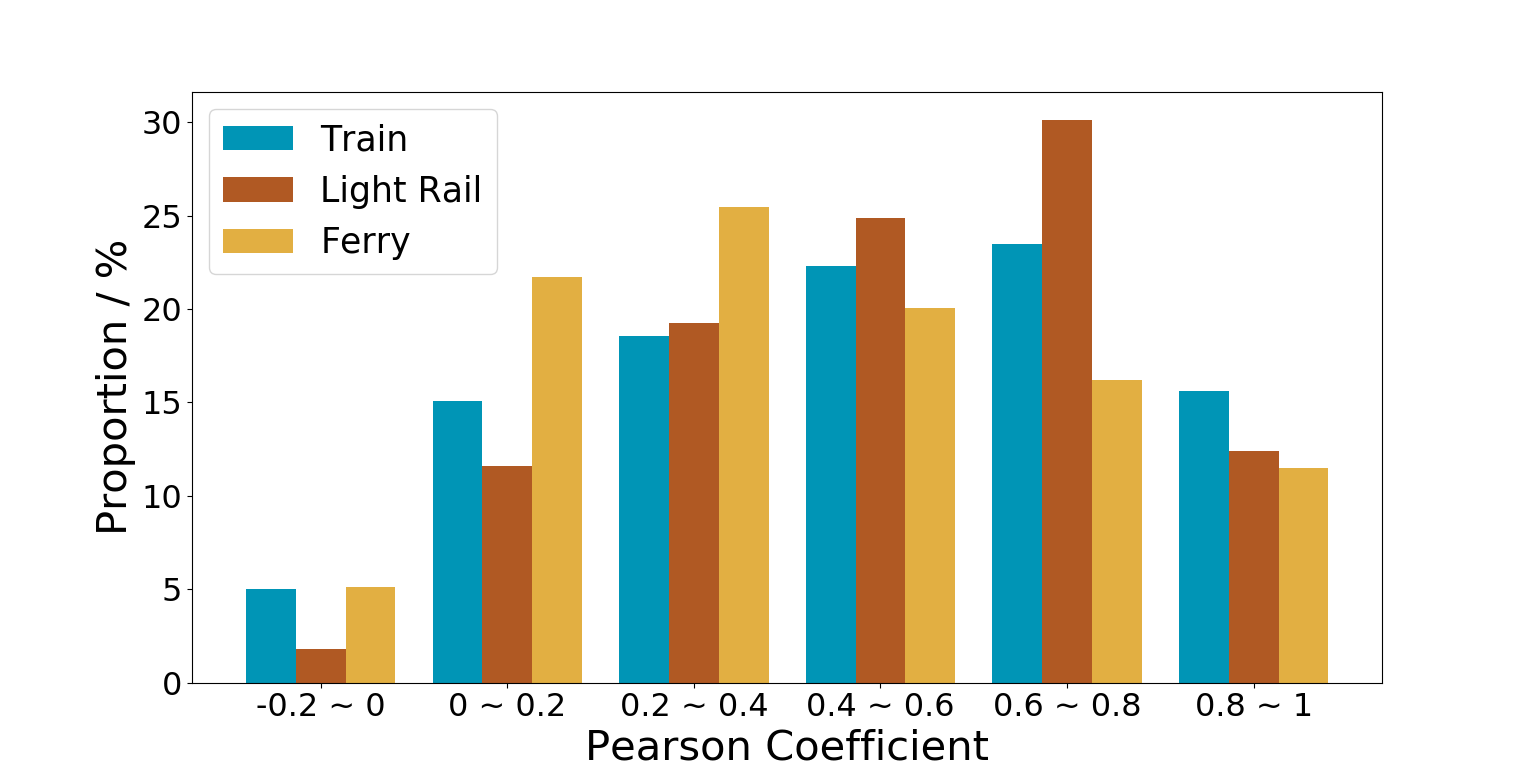}
\caption{Bus}
\label{fig:bus_sim}
\end{subfigure}
\begin{subfigure}{.48\textwidth}
\includegraphics[width=\linewidth]{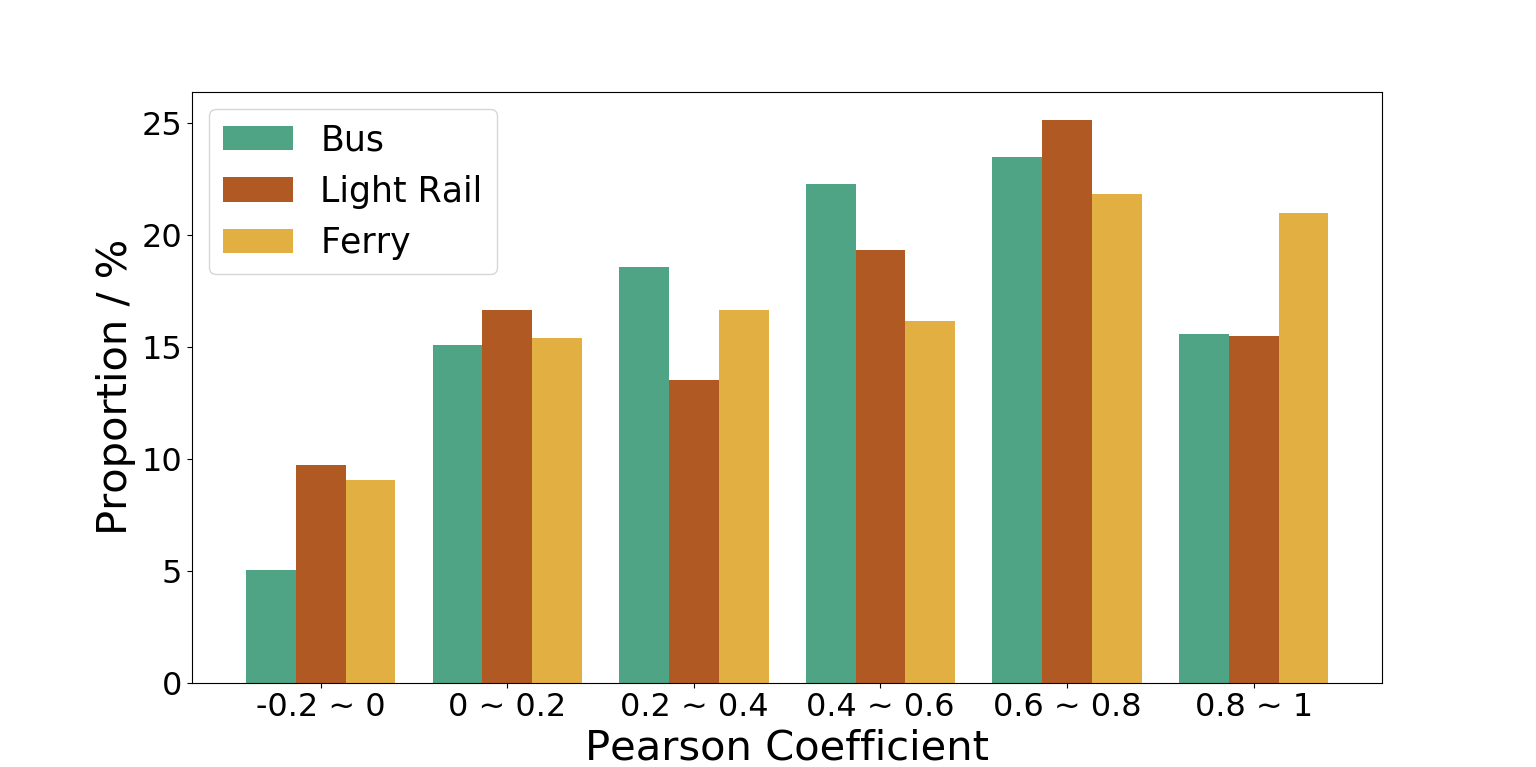}
\caption{Train}
\label{fig:train_sim}
\end{subfigure}
\begin{subfigure}{.48\textwidth}
\includegraphics[width=\linewidth]{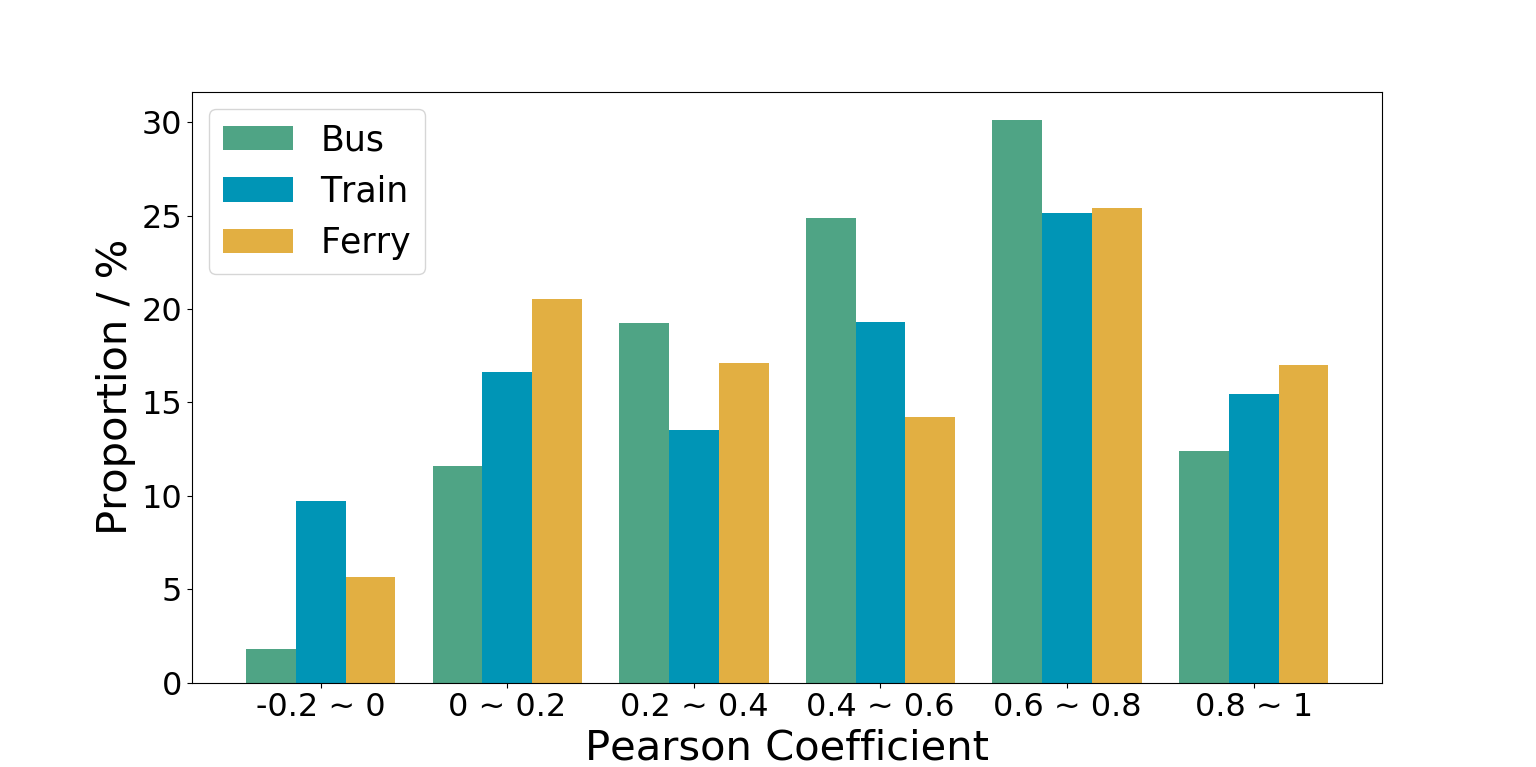}
\caption{Light Rail}
\label{fig:lr_sim}
\end{subfigure}
\begin{subfigure}{.48\textwidth}
\includegraphics[width=\linewidth]{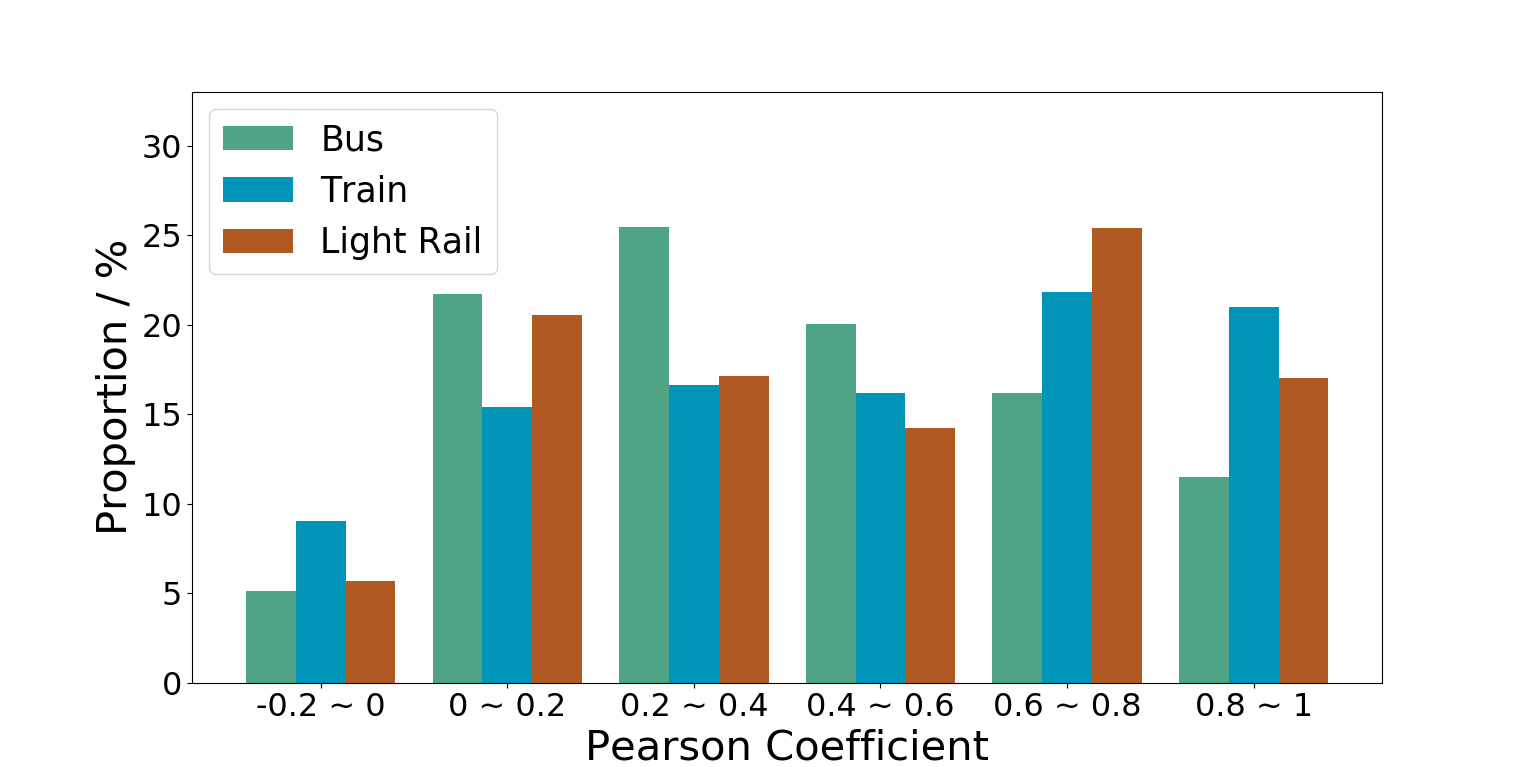}
\caption{Ferry}
\label{fig:ferry_sim}
\end{subfigure}
\caption{Correlation Analysis for Four Modes}
\label{fig:similarity}
\end{figure}

The dataset is collected from the Greater Sydney Area covering main public transport services, including buses, trains, ferries, and light rails, from 01/Apr/2017 to 30/Jun/2017. The dataset records traffic information of 24 hours a day, covering 6.37 million users. All lines’ information, including tap-on and tap-off location (e.g., name, longitude, and latitude of the station), time, and the number of passengers getting on and off, are used in experiments. The dataset does not involve personal information that can be used to identify individuals.

%\subsection{Correlation Analysis} \label{sec:correlation}

The potential correlations are visualized for four public transit modes (i.e., buses, trains, ferries, and light rails) in Sydney, where the Pearson correlation coefficient is adopted. The coefficient ranges from $-1$ to $1$, where the larger absolute value represents a higher correlation between the two modes. A positive value means a positive correlation, while a negative value means a negative correlation.

In particular, we calculate correlations between the station-based demands of different modes, and Fig.~\ref{fig:similarity} shows the distribution of coefficient values for four modes where the X-axis represents the range of coefficient values, and the Y-axis represents proportions of stations in the corresponding range. As can be seen from these figures, the proportion of stations with a correlation coefficient greater than $0.6$ is between $30\%$ and $40\%$ approximately, which illustrates the relatively wide and high correlations among any two modes.

\subsection{Demand Forecasting Problem Formulation} \label{sec:preliminary}

In this subsection, we formulate the passenger demand forecasting problem to be studied. Note that the following problem formulation is based on station-based demand. It can be readily adapted to non-station-based travel modes such as taxi and ride-sourcing service by introducing the zone-based demand concept, where ``station'' is then replaced by ``zone''.

\textbf{Demand Series.} In the multimodal transport system, we denote the demand of station $i$ at time step $t$ as a scalar $x^t_{D, i}$ for transport mode $D$ (e.g., train), which means the passenger demand between time step $t-1$ and time step $t$. Then, the demand of station $i$ over time can be represented as a vector $X_{D, i} = \{ x^1_{D, i}, x^2_{D, i}, \cdots, x^t_{D, i}, \cdots, x^T_{D, i} \}$ where T is the total number of time steps. Similarly, all stations for transport mode $D$ at time step $t$ can be represented as a vector $X_{D}^{t} = \{x_{D,1}^{t}, x_{D,2}^{t}, \cdots, x_{D,i}^{t}, \cdots, x_{D,N_{D}}^{t} \}$. Moreover, we let $\mathbf{X}_{D} = \{ X_{D}^{1}, X_{D}^{2}, \cdots, X_{D}^{t}, \cdots, X_{D}^{T} \}$ denote the demand series of transport mode $D$ over time.

\textbf{Demand Forecasting Problem with Unsupervised Knowledge Adaptation.}
Given a sequence of demand $\{ \textbf{x}_{P}^{T-\tau+1}, \cdots, \textbf{x}_{P}^{T-1}, \textbf{x}_{P}^{T} \}$ of the target mode $P$ where $\tau$ is the number of time steps utilized for prediction, the problem is defined as forecasting the demand of each station of the target mode in the future time step $T+1$:
\begin{equation}
    \hat{X}_{P}^{T+1} = \Gamma ( X_{P}^{T-\tau+1}, \cdots, X_{P}^{T-1}, X_{P}^{T})
    \label{formula:problem1}
\end{equation}
where $\Gamma(\cdot)$ is the prediction function to be learned by the model sharing network, and $\hat{X}^{T+1}_{P}$ represents the predicted demand value at time step $T+1$ of the target dataset. The above station-based formulation can be readily adapted for zone/region-based formulation if we consider zone/region-based demand rather than station-based demand.

Given the potential correlations of travel patterns between the source mode $S$ and the target mode $P$, a pre-trained recurrent model $\Phi_{s}$ learned based on the passenger demand of the source transport mode $S$ can help to enhance the forecasting performance of $P$. Thus, Eq.~\eqref{formula:problem1} can be further written as: 
\begin{equation}
    \hat{X}_{P}^{T+1} = \Gamma ( X_{P}^{T-\tau+1}, \cdots, X_{P}^{T-1}, X_{P}^{T}, \Phi{s})
\end{equation}
It should be noted that the training procedure of the target mode only takes advantage of the pre-trained model $\Phi_{s}$, and does not access the exact source dataset. Such an unsupervised knowledge adaptation strategy is different from adaptation methods in the literature, which directly utilize the source data when training the forecasting model for the target dataset.

\section{Methodology} \label{sec:model}

\begin{figure}[htb]
\centering
\begin{subfigure}{.95\textwidth}
\includegraphics[width=\linewidth]{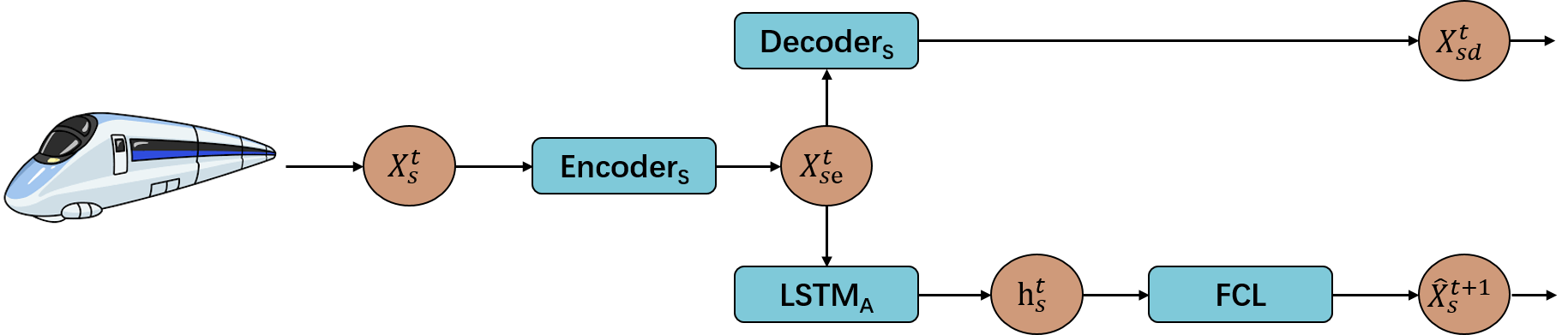}
\caption{The Procedure of Source Knowledge Extraction: $X_{S}^{t}$, $X_{Se}^{t}$, $X_{Sd}^{t}$, and $h_{S}^{t}$ denote the raw demand, the encoded vector, the decoded vector, and the hidden state at time step $t$, respectively. $h_{S}^{t}$ is the hidden state emitted by the pre-trained adaptable recurrent network $LSTM_{A}$. $\hat{X}_{S}^{t+1}$ is the predicted demand at time step $t+1$. $S$ represents the source dataset.}
\label{fig:pretrained_network}
\end{subfigure}
\begin{subfigure}{.95\textwidth}
\includegraphics[width=\linewidth]{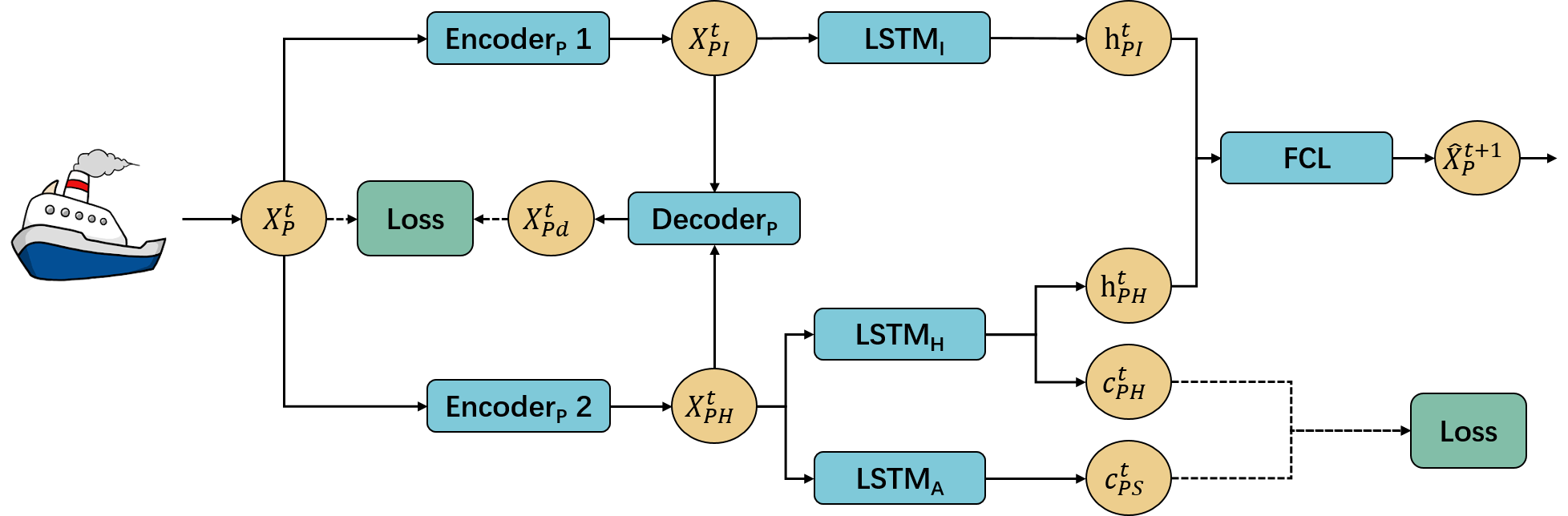}
\caption{The Structure of Model Sharing Network:  $X_{P}^{t}$, $X_{PI}^{t}$, $X_{PH}^{t}$, and $X_{Pd}^{t}$ denote the raw demand, the encoded individual knowledge part, the encoded sharing knowledge part, and the decoded vector at time step $t$, respectively. $h_{PI}^{t}$ and $h_{PH}^{t}$ are according hidden states while $c_{PI}^{t}$ and $c_{PH}^{t}$ are according memory cell emitted by the individual recurrent network $LSTM_{I}$ and sharing recurrent network $LSTM_{H}$. $c_{Ps}^{t}$ is the memory cell emitted by pre-trained adaptable recurrent network $LSTM_{A}$. And $P$ represents the target dataset.}
\label{fig:transfer_network}
\end{subfigure}
\caption{The Overall Structure of Un-Kadf}
\label{fig:model_structure}
\end{figure}

In this section, we introduce the structure and methodology of the proposed \textbf{Un-Kadf} framework as depicted in Fig.~\ref{fig:flowchart}. The first step is to deal with the demand data of the source dataset, where the temporal knowledge (not the data itself) can be recorded in the pre-trained recurrent network and further shared to the target transport mode. This is shown in Fig.~\ref{fig:pretrained_network}. Then, we elaborate on how to adapt useful knowledge from the pre-trained network to the target dataset to enhance the forecasting performance based on unsupervised knowledge adaptation. The structure of the model sharing network for unsupervised knowledge adaptation is shown in Fig.~\ref{fig:transfer_network}.

\subsection{Source Knowledge Extraction} \label{sec:pretrain}

Transfer learning consists of a two-stage learning framework, i.e., the pre-training stage to capture knowledge from source task(s), and the fine-tune stage to transfer the captured knowledge to target task(s) \citep{pan2009survey}. It provides a flexible manner to train efficiently and perform well in the computer vision (CV) area and the natural language process (NLP) area among tasks with similar distributions. As discussed in Subsection~\ref{sec:dataset}, similarities/correlations often exist among different transport modes, which can be potentially used to enhance demand forecasting. Following the success of transfer learning in CV and NLP areas to take advantage of the profitable knowledge learned from the source dataset, a pre-trained recurrent network is designed to extract and memorize the information, and an encoder-decoder framework is utilized to allow the pre-trained network to be adaptable to the target dataset. It is noteworthy that the pre-trained network rather than the source dataset is utilized for further demand forecasting, and there is no need of direct data sharing.

As shown in Fig.~\ref{fig:pretrained_network}, the raw demand data of the source dataset $X^{t}_{S} \in \mathbb{R}^{N_{S}}$ is encoded into an embedding vector $X^{t}_{Se} \in \mathbb{R}^{K}$ at first to ensure consistency between the dimensions of the source dataset and the target dataset. $N_{S}$ is the number of stations of the source dataset, and $K$ is the dimension of the encoded vector. The derivation of the LSTM layer represented in Eq.~\eqref{formula:lstm} shows that the dimension of the input vector decides the dimension of weight matrices (note that given the relevance of LSTM to this study, we briefly summarize the mechanism of LSTM in \ref{sec:LSTM}). Both datasets should have the same input dimension in order to ensure that the pre-trained recurrent network can be adapted into the target dataset. However, the number of stations of different transport modes often varies, which cannot be accommodated with the same structure of the LSTM layer. To solve such a problem, the encoder for the raw demand source data is defined as:
\begin{equation}
    X^{t}_{Se} = \tanh (\mathbf{W}_{e} X^{t}_{S} + \mathbf{b}_{e})
\end{equation}
where $\mathbf{W}_{e} \in \mathbb{R}^{K \times N_{S}}$ and $\mathbf{b}_{e}\in \mathbb{R}^{K}$ represent the weight matrix and bias vector, respectively.

Then, to extract the temporal information of the source dataset, the encoded vector is sent into the LSTM layer, following a fully connected layer to predict the demand $\hat{X}^{t+1}_{S} \in \mathbb{R}^{N_{S}}$ in the next time step. Meanwhile, in order to allow the encoded vector to keep discriminative information of the demand data, a decoder is designed to recover the raw data, which is represented as:
\begin{equation}
    X^{t}_{Sd} = \tanh (\mathbf{W}_{dS} X^{t}_{Se} + \mathbf{b}_{dS})
\end{equation}
where $\mathbf{W}_{dS} \in \mathbb{R}^{N_{S} \times K}$ and $\mathbf{b}_{dS}\in \mathbb{R}^{N_{S}}$ represent the weight matrix and bias vector, respectively.

To ensure that the information recorded in the pre-trained network can describe patterns of the source dataset more comprehensively, we need to optimize the network to avoid adapting noises or useless knowledge \citep{pan2009survey}. In the training process of the source dataset, the objective is to minimize the error between the real demand and the predicted values and the error between the real demand and recovered demand values. The loss function is defined as the mean squared error for time step length $\tau$, which is formulated as follows:
\begin{equation}
   L(\theta) = \sum^{T+\tau}_{t=T+1}|| \hat{X}_{S}^{t} - X_{S}^{t} || + \sum^{T+\tau-1}_{t=T}|| X_{Sd}^{t} - X_{S}^{t} ||
\label{formula:loss1}
\end{equation}
where $\hat{X}^{T+1}_{S}$ represents the predicted demand value at time step $T+1$ of the source dataset and $\theta$ denotes all the learnable parameters in the pre-trained network. And the network is trained via back-propagation and the Adam optimizer.

\subsection{Model Sharing Network} \label{sec:sharing}

As revealed in Subsection~\ref{sec:dataset}, two transport modes may share some similarities and hold disparities at the same time. Therefore, the raw demand data of the target dataset is encoded into two parts, i.e., the encoded individual knowledge part $X^{t}_{PI} \in \mathbb{R}^{K}$ and the encoded sharing knowledge part $X^{t}_{PH} \in \mathbb{R}^{K}$ which are calculated as:
\begin{equation}
\begin{aligned}
    & X^{t}_{PI} = \tanh (\mathbf{W}_{PI} X^{t}_{P} + \mathbf{b}_{PI}) \\
    & X^{t}_{PH} = \tanh (\mathbf{W}_{PH} X^{t}_{P} + \mathbf{b}_{PH})
\end{aligned}
\end{equation}
where $\mathbf{W}_{PI} \in \mathbb{R}^{K \times N_{S} }$, $\mathbf{W}_{PH} \in \mathbb{R}^{K \times N_{S}}$ are learnable weight matrices and $\mathbf{b}_{dS}\in \mathbb{R}^{K}$ and $\mathbf{b}_{dS}\in \mathbb{R}^{K}$ are bias vectors.

Then, to analyze the temporal knowledge of the individual and sharing parts independently, two different LSTM networks are applied to get hidden states and internal memory cells as:
\begin{equation}
\begin{aligned}
    & \mathbf{h}^{t}_{PI}, \mathbf{c}^{t}_{PI} = LSTM_{I} (X^{t}_{PI}, \mathbf{h}^{t-1}_{PI}, \mathbf{c}^{t-1}_{PI}) \\
    & \mathbf{h}^{t}_{PH}, \mathbf{c}^{t}_{PH} = LSTM_{H} (X^{t}_{PH}, \mathbf{h}^{t-1}_{PH}, \mathbf{c}^{t-1}_{PH})
\end{aligned}
\end{equation}
The following fully connected layer is used to combine hidden states $\mathbf{h}^{t}_{PI}$ and $\mathbf{h}^{t}_{PH}$to get the predicted demand value of the target dataset as:
\begin{equation}
    \hat{X}^{t+1}_{P} = \tanh (\mathbf{W}_{P} (\mathbf{h}^{t}_{PI} + \mathbf{h}^{t}_{PH}) + \mathbf{b}_{P})
\end{equation}
where $\mathbf{W}_{P} \in \mathbb{R}^{N_{S} \times 2K}$ and $\mathbf{b}_{P}\in \mathbb{R}^{N_{S}}$ denote the weight matrix and bias vector, respectively.

Furthermore, as introduced in Subsection~\ref{sec:pretrain}, the pre-trained network memorizes travel patterns of the source dataset. Therefore, motivated by the success of transfer learning for trajectory prediction and precipitation nowcasting \citep{yao2020unsupervised}, we adopt such a strategy for unsupervised knowledge adaptation. In detail, to make the sharing recurrent network $LSTM_{H}$ be able to represent the sharing knowledge of both datasets, the sharing knowledge part is also sent into the adaptable pre-trained recurrent network $LSTM_{A}$ to get the internal memory cell $\mathbf{c}^{t}_{PS}$. The values of $\mathbf{c}^{t}_{PS}$ and $\mathbf{c}^{t}_{PH}$ are similar, which means that the parameters (e,g., weight matrices and bias vectors) of the pre-trained network and the sharing network are similar. The parameters of the pre-trained recurrent network $LSTM_{A}$ are frozen during the training process so that the sharing network can focus on more sharing patterns.

Similar to the decoder in the pre-trained recurrent network, encoded individual and sharing parts are recovered by the decoder to improve the representation ability of the encoded vectors as:
\begin{equation}
    X^{t}_{Pd} = \tanh (\mathbf{W}_{dP} (X^{t}_{PI} + X^{t}_{PH}) + \mathbf{b}_{dP})
\end{equation}
where $\mathbf{W}_{dP} \in \mathbb{R}^{N_{S} \times 2K}$ and $\mathbf{b}_{dP}\in \mathbb{R}^{N_{S}}$ are the weight matrix and bias vector, respectively.

\subsection{Training Strategy} \label{sec:training}

In the training process, one objective is to minimize the error between the values of internal memory cells obtained from the sharing recurrent network and pre-trained recurrent network to learn sharing travel patterns. Moreover, to enhance the forecasting performance of the target dataset in the training process of the model sharing recurrent network, the error between the true demand and the predicted values, and the error between the real demand and recovered demand data are needed to be minimized. The loss function is defined as the mean squared error for time step length $\tau$, which is represented as:
\begin{equation}
\begin{aligned}
    & L1 =  \sum^{T+\tau}_{t=T+1}|| \hat{X}_{P}^{t} - X_{P}^{t} || \\
    & L2 = \sum^{T+\tau-1}_{t=T}|| X_{Pd}^{t} - X_{P}^{t} || \\
    & L3 = \sum^{T+\tau-1}_{t=T}|| c_{PH}^{t} - c_{PS}^{t} || \\
    & L(\Pi) =  L1 + \gamma \times L2 + \beta \times L3
\end{aligned}
\label{formula:loss2}
\end{equation}
where $\Pi$ denotes all the learnable parameters in the model sharing network. $\gamma$ and $\beta$ are hyper-parameters to decide the weights of data recovery loss $L2$ and memory update loss $L3$, which will be evaluated in Subsection~\ref{sec:hyper}. And the network is trained via back-propagation and the Adam optimizer. Note the training procedures of the knowledge adaptation framework are unsupervised, which does need the exact source data for optimization as discussed in Section~\ref{sec: introduction}.

\section{Experiments} \label{sec:experiment}

This section presents results from experiments carried out in this study. The experimental setup, including dataset setting, evaluation matrices, and network implementation, is presented at first. The components of the proposed model and hyper-parameter sensitivity are then discussed and analyzed. Furthermore, we compare the proposed model with a list of baseline models and state-of-the-art methods. 

\subsection{Experimental Setup}

\textbf{Dataset Setting.} 
The demand data is normalized by Min-Max normalization for training and re-scaled to the actual values for evaluating the prediction performance. To test the performance of \textbf{Un-Kadf}, 60\% data are used for training, 20\% data are used for validation, and the rest for testing. In each experiment, we use the data of one transport mode as the source dataset and the data of another mode as the target dataset. The time step length we choose in the experiments is one hour. Since the data volume is too large, which contains some meaningless data (e.g., the demand is zero for more than $80\%$ of time steps in one day), we drop the stations with zero demand per hour for more than $60\%$ of time steps. The number of stations for the bus, train, light rail, and ferry stations is 472, 250, 23, and 33, respectively. And we use the previous 12 time steps (12 hours) to predict the public transport demand in the next time step (next one hour).

\textbf{Evaluation Matrices.} 
Three evaluation matrices are used to evaluate the proposed model: Root Mean Square Error (RMSE), Mean Absolute Error (MAE), and Mean Absolute Percentage Error (MAPE).

\textbf{Network Implementation.}
The batch size is set to 64 and the learning rate is fixed as $0.0001$. $1000$ episodes are run for model training. The proposed model is tuned with the hyper-parameters $\gamma$ and $\beta$ in the loss function Eq.~\eqref{formula:loss2} (from 0.1 to 1.0 with a step size of 0.1). The number of hidden states is fixed as $64$ of encoder layers and LSTM layers for pre-trained model adaptation. Different hyper-parameter values for different transit modes are set, which are shown in Table~\ref{table:sensitivity}, and the details of hyper-parameter tuning will be introduced in Subsection~\ref{sec:hyper}. 

\begin{table}[htb]
\small
\caption{Hyper-parameter Setting of \textbf{Un-Kadf}}
\setlength{\tabcolsep}{.8mm}{
\begin{tabular}{c|c|c|c|c|c|c|c|c|c|c|c|c}
\hline
\textbf{Mode} & \multicolumn{6}{c|}{\textbf{Value of $\gamma$}} & \multicolumn{6}{c}{\textbf{Value of $\beta$}} \\ \hline
\textbf{Bus} & \textbf{Train} & 0.4 & \textbf{Light Rail} & 1 & \textbf{Ferry} & 0.5 & \textbf{Train} & 1 & \textbf{Light Rail} & 0.4 & \textbf{Ferry} & 0.1 \\
\textbf{Train} & \textbf{Bus} & 0.6 & \textbf{Light Rail} & 0.6 & \textbf{Ferry} & 0.4 & \textbf{Bus} & 0.7 & \textbf{Light Rail} & 0.9 & \textbf{Ferry} & 0.7 \\
\textbf{Light Rail} & \textbf{Bus} & 0.9 & \textbf{Train} & 0.3 & \textbf{Ferry} & 1.0 & \textbf{Bus} & 0.3 & \textbf{Train} & 0.1 & \textbf{Ferry} & 0.6 \\
\textbf{Ferry} & \textbf{Bus} & 0.1 & \textbf{Train} & 0.6 & \textbf{Light Rail} & 0.5 & \textbf{Bus} & 0.6 & \textbf{Train} & 0.6 & \textbf{Light Rail} & 0.6 \\ \hline
\end{tabular}}
\label{table:sensitivity}
\end{table}

\subsection{Network Architecture Analysis}

To provide a more concrete understanding of the effectiveness of the encoder-decoder framework and the unsupervised knowledge adaptation strategy described in Section~\ref{sec:model}, this subsection compares the proposed method against several variants as described below: 

\begin{itemize}
    \item \textbf{LSTM}: The LSTM layer is directly used to model long-and-short-term dependencies, followed by a fully connected layer to predict the demand for each transport mode independently.
    \item \textbf{Encoder-Adaptation}: The encoded vectors of the target dataset are only sent into LSTM layers for temporal knowledge extraction. The decoder to recover the raw demand data is removed in this architecture.
    \item \textbf{Encoder-Decoder}: The knowledge adaptation from the source dataset to the target dataset is removed, which means that only encoder and decoder components are used to help LSTM for features capturing and further demand prediction.
    \item \textbf{Encoder-LSTM}: This architecture removes both the decoder and adaptation module. The raw demand data is sent into encoders and LSTM layers for further prediction.
\end{itemize}

\begin{table}[htb]
\caption{Network Architecture Analysis with Different variants}
\begin{tabular}{c|c|cccc|c}
\hline
\textbf{\begin{tabular}[c]{@{}c@{}}Evaluation \\ Matrices\end{tabular}} & \textbf{Mode} & \textbf{LSTM} & \textbf{\begin{tabular}[c]{@{}c@{}}Encoder-\\Adaptation\end{tabular}} & \textbf{\begin{tabular}[c]{@{}c@{}}Encoder-\\Decoder\end{tabular}} & \textbf{\begin{tabular}[c]{@{}c@{}}Encoder-\\LSTM\end{tabular}} & \textbf{Un-Kadf} \\ \hline
\multirow{4}{*}{\textbf{MAE}} & \textbf{Bus} & 8.750 & 8.233 & 8.397 & 8.545 & \textbf{7.841} \\
 & \textbf{Train} & 24.357& 22.810 & 21.863 & 24.071 & \textbf{19.614} \\
 & \textbf{Light Rail} & 12.106 & 10.863 & 10.908 & 12.253 & \textbf{10.530} \\
 & \textbf{Ferry} & 14.298 & 13.924 & 14.232 & 14.361 & \textbf{13.039} \\ \hline
\multirow{4}{*}{\textbf{RMSE}} & \textbf{Bus} & 20.108 & 18.824 & 19.274 & 19.832 & \textbf{17.814} \\
 & \textbf{Train} & 70.262 & 65.662 & 63.635 & 69.422 & \textbf{54.693} \\
 & \textbf{Light Rail} & 25.819 & 21.899 & 22.166 & 25.144 & \textbf{21.381} \\
 & \textbf{Ferry} & 37.356 & 38.745 & 38.788 & 38.941 & \textbf{35.832} \\ \hline
\multirow{4}{*}{\textbf{MAPE}} & \textbf{Bus} & 0.169 & 0.155 & 0.157 & 0.168 & \textbf{0.149} \\
 & \textbf{Train} & 0.157 & 0.151 & 0.146 & 0.158 & \textbf{0.130} \\
 & \textbf{Light Rail} & 0.180 & 0.165 & 0.168 & 0.183 & \textbf{0.160} \\
 & \textbf{Ferry} & 0.210 & 0.192 & 0.199 & 0.206 & \textbf{0.187} \\ \hline
\end{tabular}
\label{table:ablation}
\end{table}

The forecasting results under different architectures are listed in Table~\ref{table:ablation}. Since Encoder-Adaptation and \textbf{Un-Kadf} deal with two transport modes (i.e., the source mode and target mod), the results of them are calculated as average values of the forecasting results for each mode. Several observations are summarized as follows based on the experiments of network architectures:

First, the forecasting results of LSTM and Encoder-LSTM are similar, which do not show a significant difference. These results illustrate that the design of the independent encoder to split the target dataset into individual and sharing knowledge parts does not destroy the travel patterns of demand data. It illustrates the rationality of the encoder introduced in Subsection~\ref{sec:sharing}. Second, Encoder-LSTM and LSTM yield higher MAE, RMSE, and MAPE than Encoder-Decoder, which means that the recovery procedure operated by the decoder is able to enhance the representation ability of the encoded individual and sharing knowledge vectors of the target dataset. Besides, when compared with Encoder-LSTM and LSTM,  Encoder-Adaptation achieves higher accuracy for four transit modes. This highlights the usefulness of adapting the knowledge recorded in the pre-trained recurrent network optimized using the source dataset does improve the prediction performance of the target dataset. Furthermore, the overall model \textbf{Un-Kadf} gains lower values of MAE, RMSE, and MAPE than other listed architectures. In view of the above, the encoder-decoder framework that formulates encoded vectors and the unsupervised knowledge adaptation strategy to adapt the learned knowledge from the source dataset to the target dataset can both help improve the forecasting performance.

\subsection{Hyper-parameter Sensitivity} \label{sec:hyper}

The sensitivity of hyper-parameters $\gamma$ and $\beta$ in the loss function Eq.~\eqref{formula:loss2} to decide the weights of the data recovery loss and the memory update loss for \textbf{Un-Kadf} training are studied in this subsection.

\begin{figure}[htb]
\centering
\begin{subfigure}{.48\textwidth}
\includegraphics[width=\linewidth]{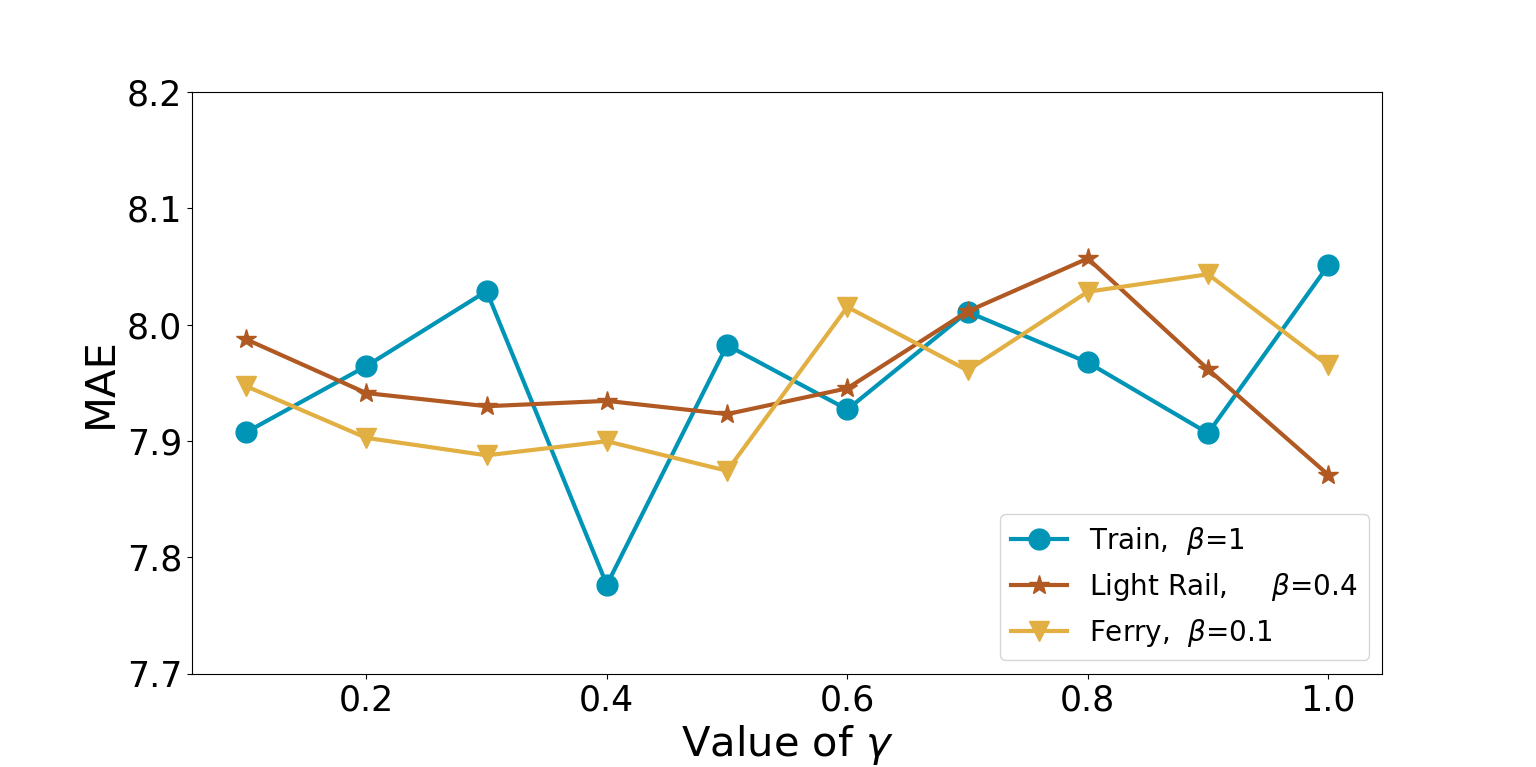}
\caption{Bus}
\label{fig:gamma_bus}
\end{subfigure}
\begin{subfigure}{.48\textwidth}
\includegraphics[width=\linewidth]{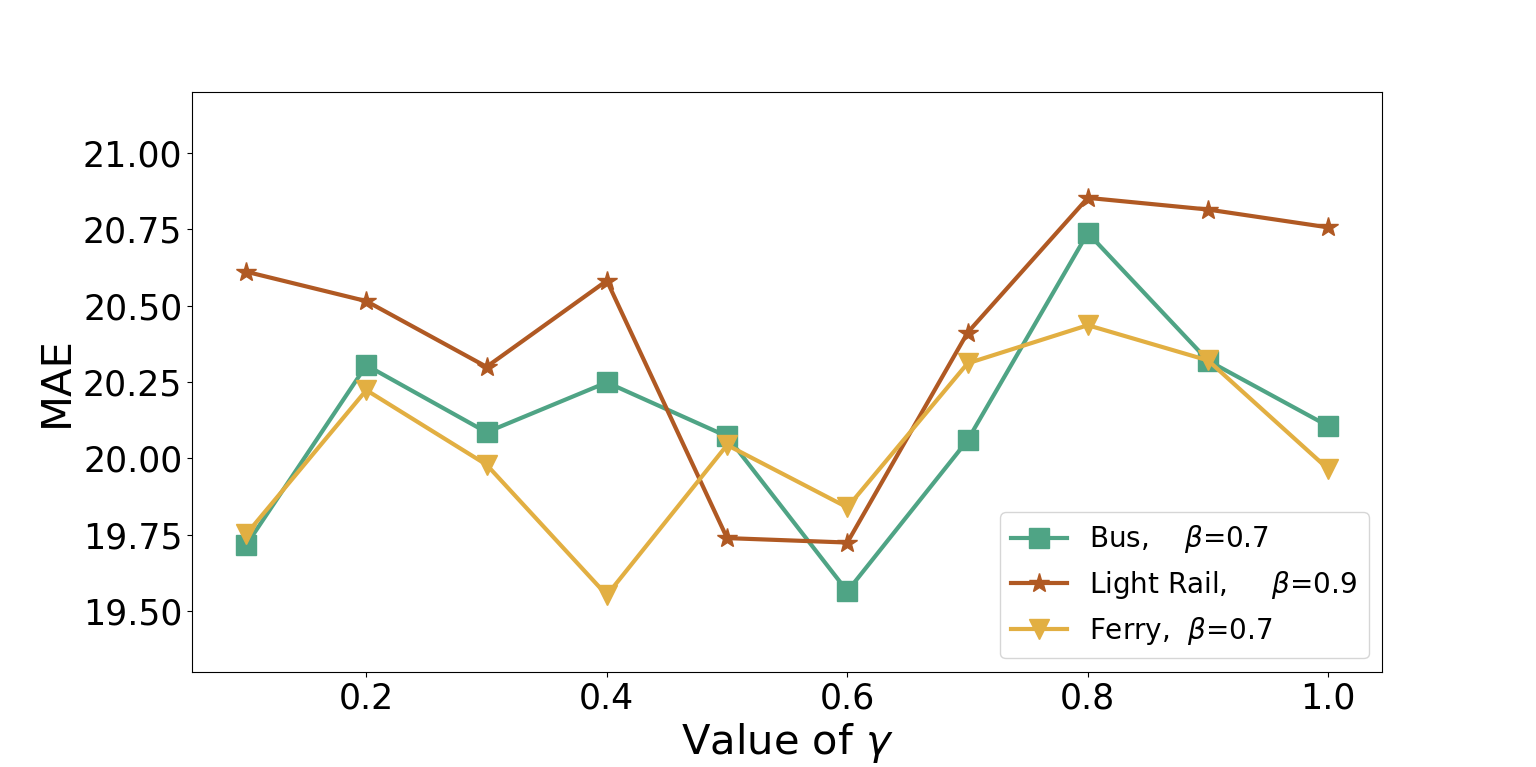}
\caption{Train}
\label{fig:gamma_train}
\end{subfigure}
\begin{subfigure}{.48\textwidth}
\includegraphics[width=\linewidth]{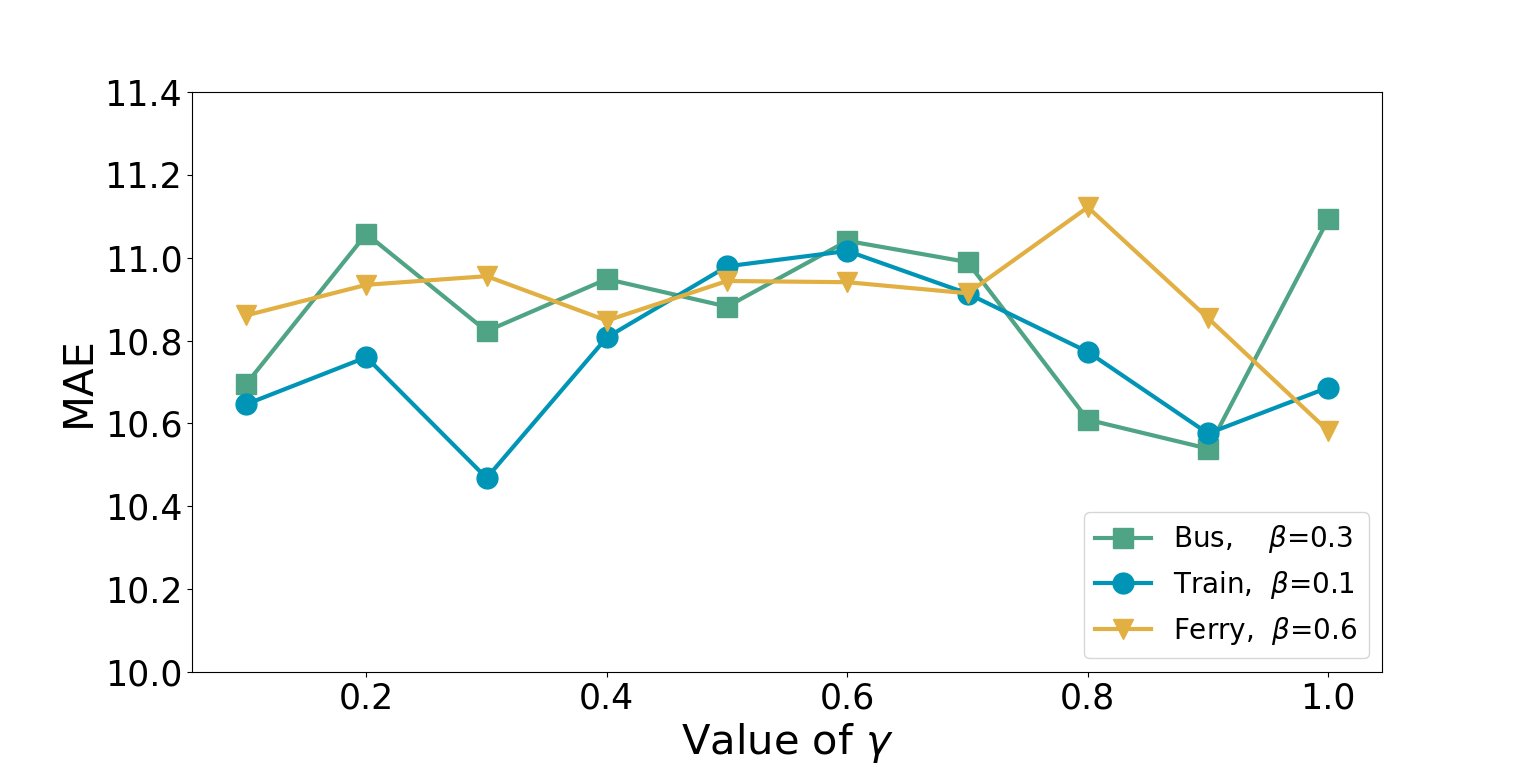}
\caption{Light Rail}
\label{fig:gamma_lr}
\end{subfigure}
\begin{subfigure}{.48\textwidth}
\includegraphics[width=\linewidth]{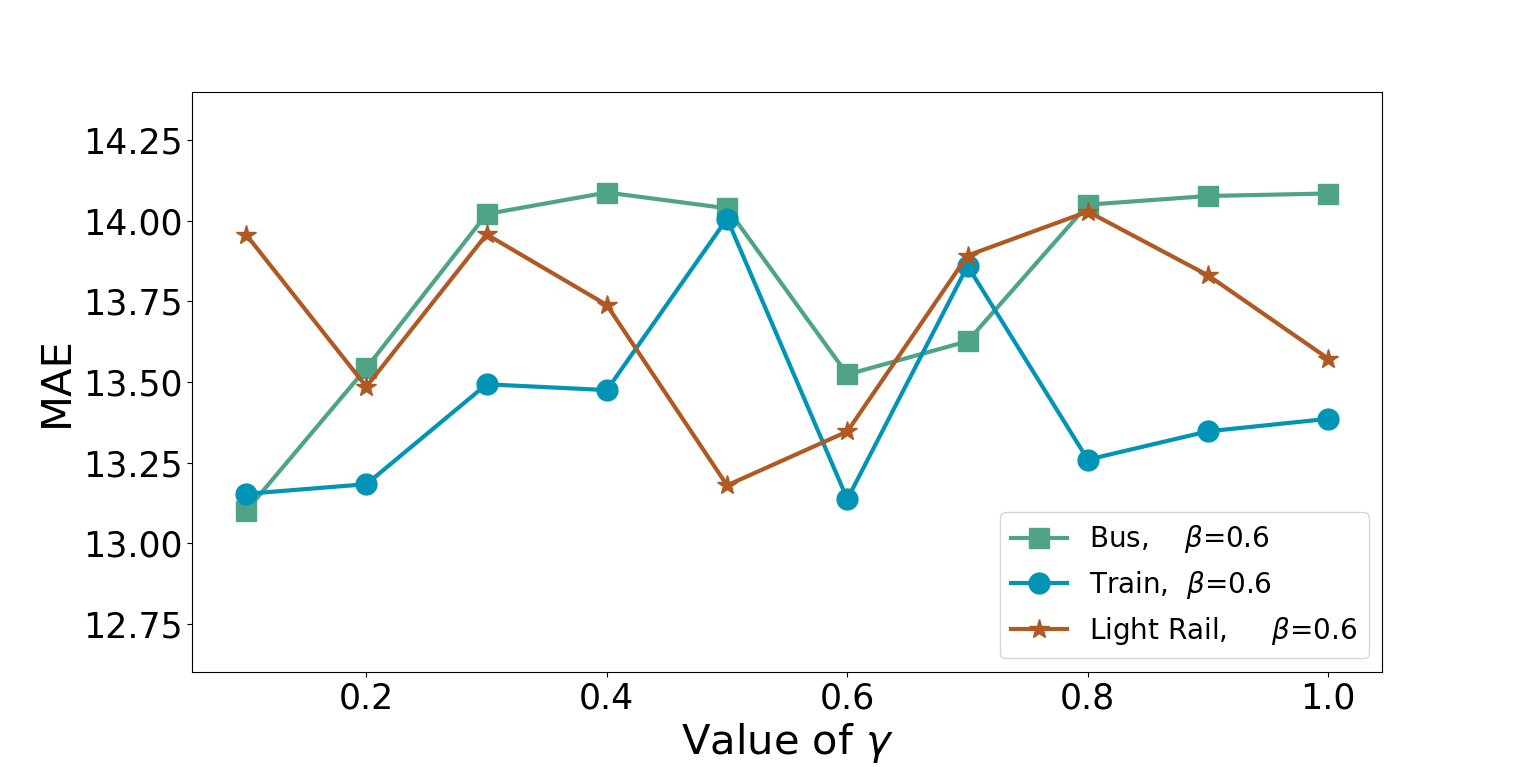}
\caption{Ferry}
\label{fig:gamma_ferry}
\end{subfigure}
\caption{Sensitivity Analysis of Hyper-parameter $\gamma$}
\label{fig:gamma}
\end{figure}

\begin{figure}[htb]
\centering
\begin{subfigure}{.48\textwidth}
\includegraphics[width=\linewidth]{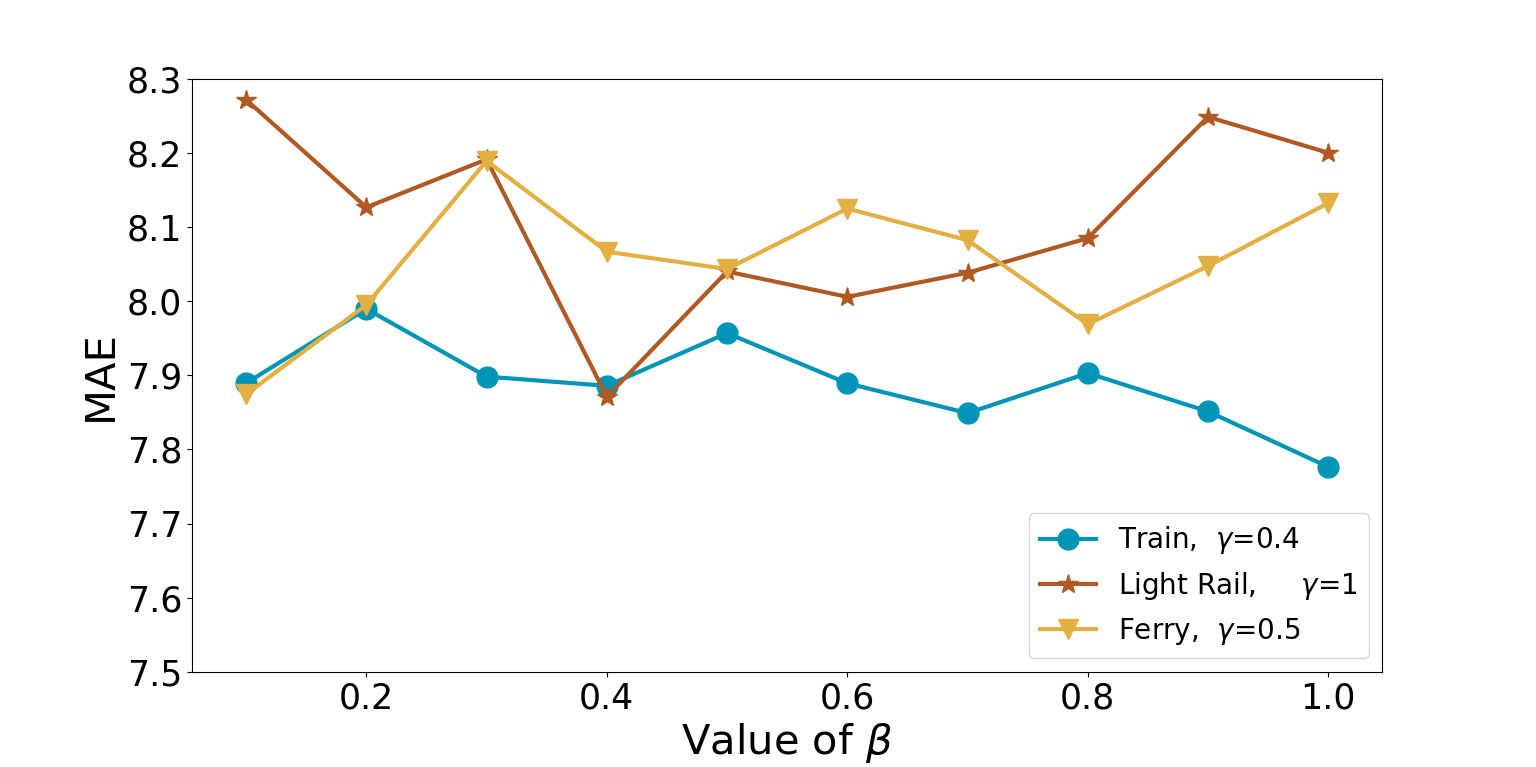}
\caption{Bus}
\label{fig:beta_bus}
\end{subfigure}
\begin{subfigure}{.48\textwidth}
\includegraphics[width=\linewidth]{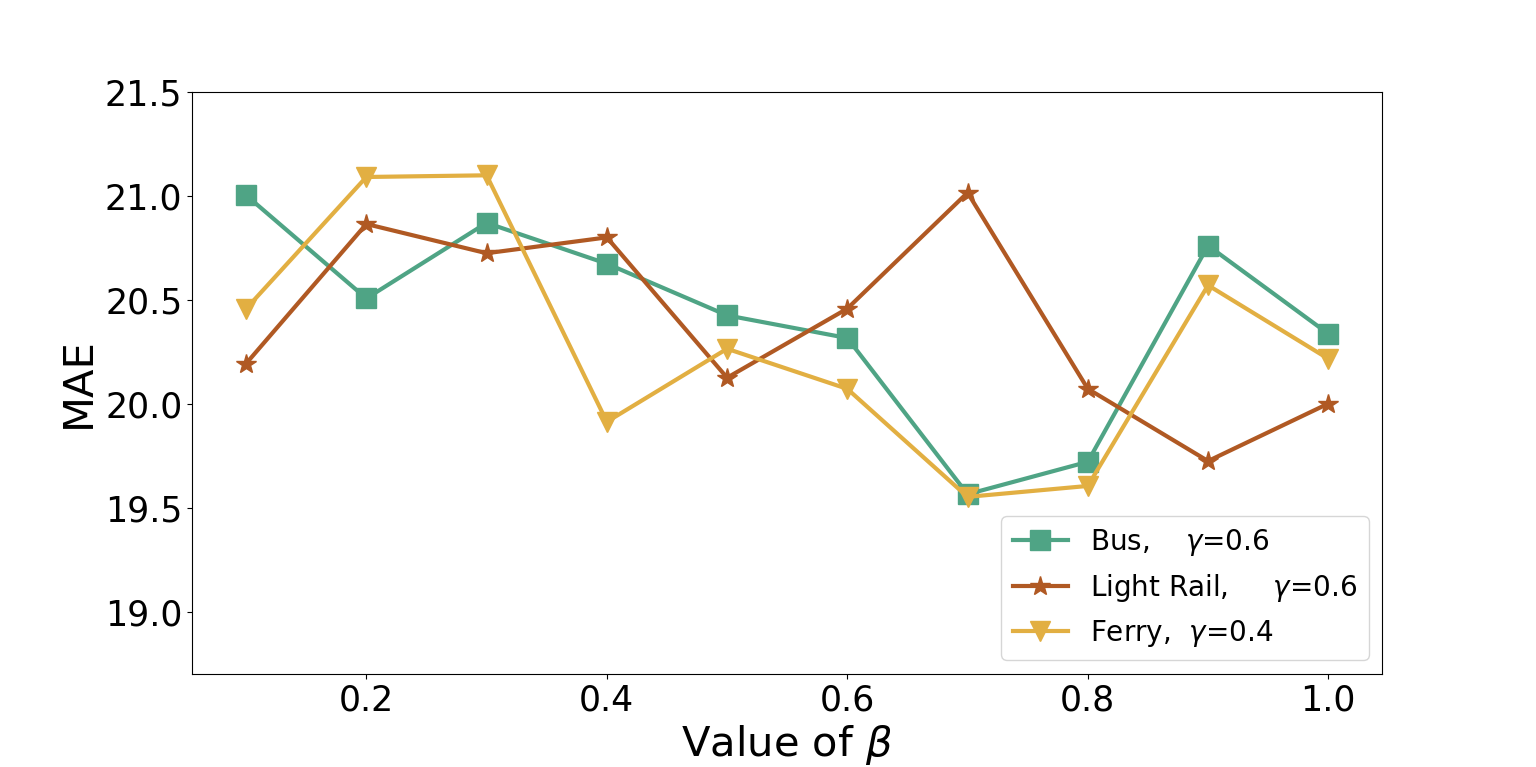}
\caption{Train}
\label{fig:beta_train}
\end{subfigure}
\begin{subfigure}{.48\textwidth}
\includegraphics[width=\linewidth]{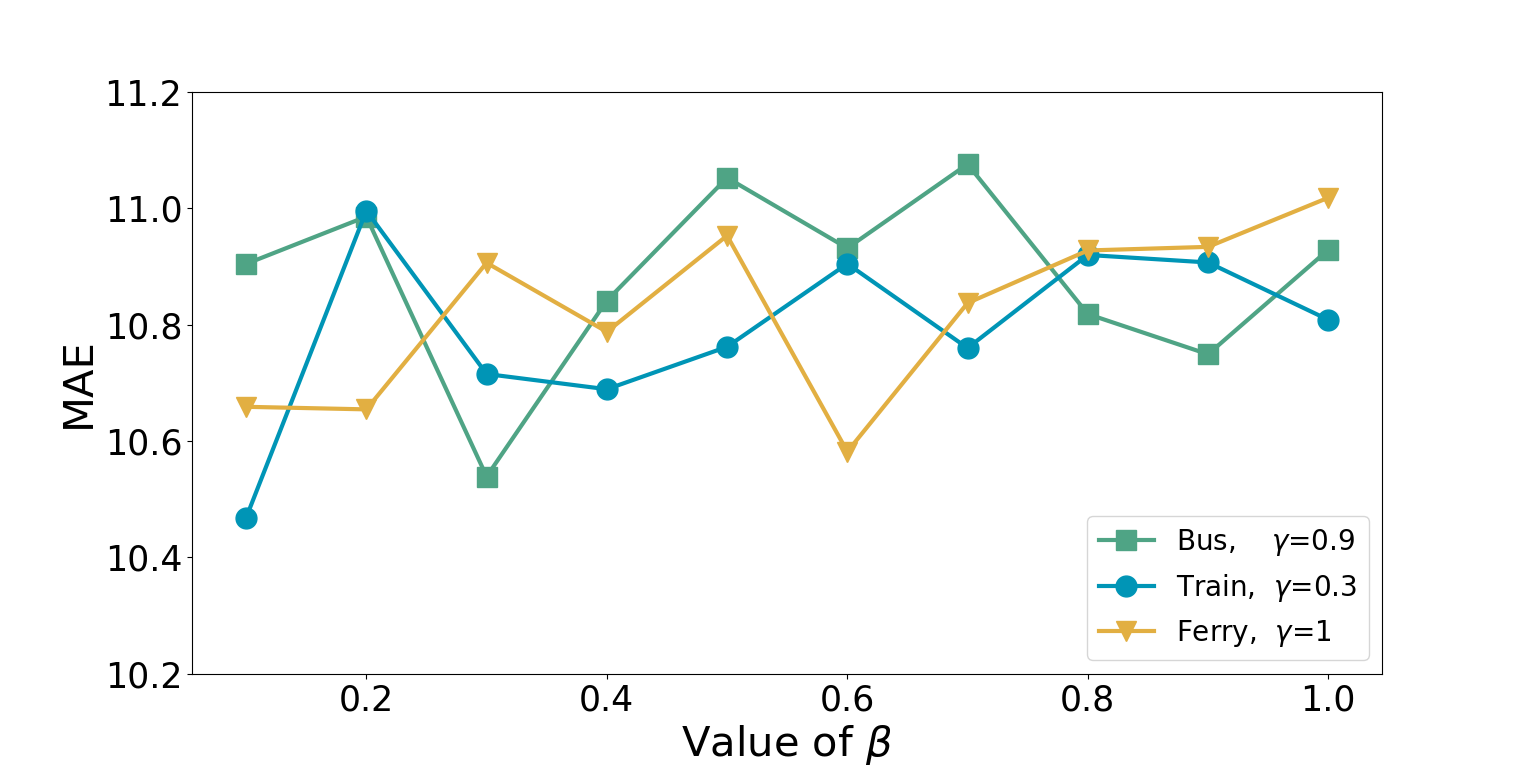}
\caption{Light Rail}
\label{fig:beta_lr}
\end{subfigure}
\begin{subfigure}{.48\textwidth}
\includegraphics[width=\linewidth]{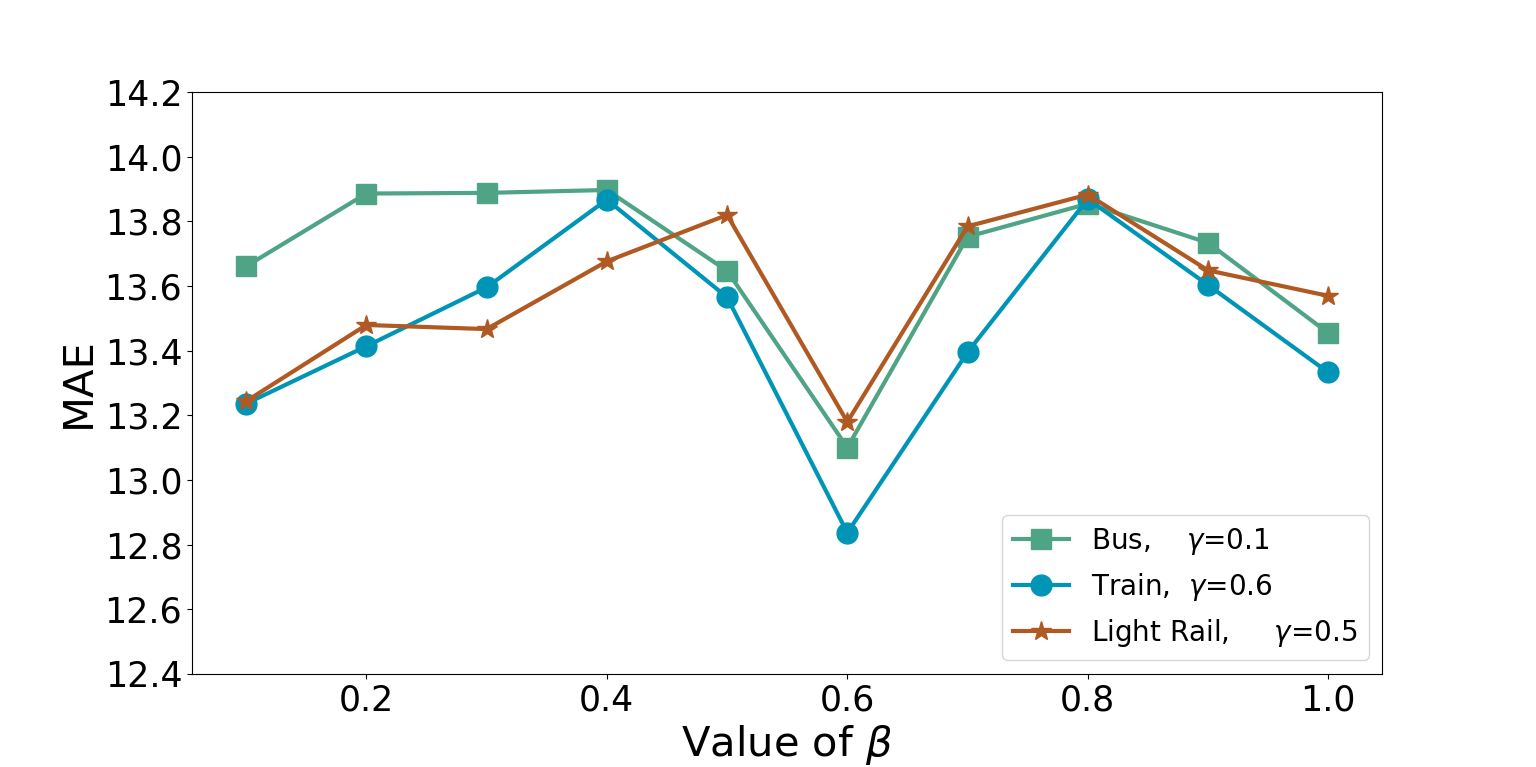}
\caption{Ferry}
\label{fig:beta_ferry}
\end{subfigure}
\caption{Sensitivity Analysis of Hyper-parameter $\beta$}
\label{fig:beta}
\end{figure}

\begin{table}[htb]
\caption{Forecasting Performance of Special Hyper-parameter Setting}
\setlength{\tabcolsep}{.6mm}{
\begin{tabular}{c|ccc|ccc|ccc}
\hline
\textbf{Hyper-parameter} & \multicolumn{3}{c|}{\textbf{$\gamma$=0}} & \multicolumn{3}{c|}{\textbf{$\beta$=0}} & \multicolumn{3}{c}{\textbf{$\gamma$=$\beta$=0}} \\ \hline
\textbf{Mode} & \textbf{MAE} & \textbf{RMSE} & \textbf{MAPE} & \textbf{MAE} & \textbf{RMSE} & \textbf{MAPE} & \multicolumn{1}{c}{\textbf{MAE}} & \multicolumn{1}{c}{\textbf{RMSE}} & \multicolumn{1}{c}{\textbf{MAPE}} \\ \hline
\textbf{Bus} & 8.211 & 18.908 & 0.155 & 8.357 & 18.961 & 0.156 & 8.573 & 20.112 & 0.172 \\
\textbf{Train} & 22.143 & 64.505 & 0.149 & 21.649 & 63.968 & 0.143 & 23.569 & 68.692 & 0.156 \\
\textbf{Light Rail} & 10.899 & 22.219 & 0.164 & 10.849 & 22.060 & 0.164 & 12.750 & 25.278 & 0.184 \\
\textbf{Ferry} & 13.781 & 37.870 & 0.191 & 14.163 & 38.409 & 0.195 & 14.364 & 39.646 & 0.208 \\ \hline
\end{tabular}}
\label{table:betagamma}
\end{table}

Different settings of hyper-parameters are tested for the proposed model based on the validation set. The setting of hyper-parameters that yields the best performance is then utilized to conduct testing based on the testing set, which has been listed in Table~\ref{table:sensitivity}. Due to the large number of experiments conducted for hyper-parameters (i.e., $10 \times 10 \times 3 = 300$ experiments conducted for each transport mode), we only show MAE for two groups of experimental results in the manuscript, i.e., the performance with $\gamma$ changing from $0.1$ to $1.0$ and $\beta$ fixed as the value yields the best performance for each mode is shown in Fig.~\ref{fig:gamma}, and the performance with $\beta$ changing from $0.1$ to $1.0$ and $\gamma$ fixed as the value yields the best performance for each mode is presented in Fig.~\ref{fig:beta}. As can be seen from these figures, MAE does not change drastically when the hyper-parameters change, which illustrates the robustness of the proposed model with respect to these hyper-parameters. To further illustrate this, we calculate the standard deviation of the listed MAE in the aforementioned figures. When fixing $\beta$, the values of standard deviation are $0.062$ for the bus, $0.365$ for the train, $0.172$ for the light rail, and $0.336$ for the ferry. When fixing $\gamma$, the values of standard deviation are $0.129$ for the bus, $0.462$ for the train, $0.150$ for the light rail, and $0.267$ for the ferry. These stationary experimental results illustrate the stability of \textbf{Un-Kadf}, which is not sensitive to the settings of hyper-parameters.

Furthermore, it is noteworthy that when the value of $\gamma$ is set to zero, the decoder module to recover the raw demand data becomes invalid, which becomes similar to the Encoder-Adaptation structure. And when the value of $\beta$ becomes zero, the knowledge learned from the source dataset recorded in the pre-trained network can not adapt to the target dataset for forecasting enhancement, which becomes similar to the Encoder-Decoder structure. Also, if $\gamma$ and $\beta$ are zero, the network becomes similar to Encoder-LSTM. The results (i.e., the average values of the forecasting results for each mode) of these three architectures are listed in Table~\ref{table:betagamma}, which is consistent with the performance shown in Table~\ref{table:ablation}. Their less satisfactory performance demonstrates the effectiveness of the encoder-decoder framework and unsupervised knowledge adaptation via model sharing.

\subsection{Overall Comparison}

In this subsection, we compare the proposed model with several baseline models and state-of-the-art methods, which are summarized in the following.

\begin{table}[h]
\caption{Hyper-parameter Settings of Compared Models with Single-Mode}
\setlength{\tabcolsep}{.6mm}{
\begin{tabular}{c|c|cccc}
\hline
\textbf{Model} & \textbf{Hyper-parameter} & \textbf{Bus} & \textbf{Train} & \textbf{Light Rail} & \textbf{Ferry} \\ \hline
\multirow{2}{*}{\textbf{MLP}} & \textbf{No. of Layers} & 2 & 2 & 2 & 2 \\
 & \textbf{No. of Units of Layers} & {512, 256} & {128, 128} & {128, 64} & {64, 64} \\ \hline
\multirow{2}{*}{\textbf{LSTM}} & \textbf{No. of Layers} & 2 & 2 & 2 & 2 \\
 & \textbf{No. of Units of Layers} & {128, 64} & {64, 64} & {64, 64} & {64, 64} \\ \hline
\textbf{DA-RNN} & \textbf{No. of Units of Layers} & {256, 128} & {256, 32} & {32, 32} & {256, 128} \\ \hline
\multirow{2}{*}{\textbf{ConvLSTM}} & \textbf{No. of Units of Layers} & {64, 32} & {64, 32} & {32, 32} & {128, 32} \\
 & \textbf{Kernel Size} & 5 & 5 & 5 & 5 \\ \hline
\multirow{3}{*}{\textbf{Transformer}} & \textbf{No. of Encoder Layers} & 1 & 3 & 3 & 3 \\
 & \textbf{\begin{tabular}[c]{@{}c@{}} No. of heads for \\ multihead-attention models \end{tabular}} & 8 & 4 & 6 & 8 \\
 & \textbf{No. of Units of Layer} & 64 & 128 & 128 & 128 \\ \hline
\multirow{3}{*}{\textbf{DSANet}} & \textbf{No. of Units of Recurrent Layer} & 64 & 64 & 128 & 32 \\
 & \textbf{No. of Units of Convolution Layer} & 64 & 32 & 64 & 32 \\
 & \textbf{Filter Length} & 3 & 3 & 3 & 5 \\ \hline
\end{tabular}}
\label{table: paramter1}
\end{table}

\begin{table}[h]
\caption{Hyper-parameter Settings of AK2M2}
\setlength{\tabcolsep}{0.4mm}{
\begin{tabular}{c|c|c|c|c|c|c}
\hline
\textbf{Mode} & \textbf{\begin{tabular}[c]{@{}c@{}}Cooperative \\ Mode\end{tabular}} & \textbf{\begin{tabular}[c]{@{}c@{}}Epsilon in \\ Loss Function\end{tabular}} & \textbf{Gamma} & \textbf{\begin{tabular}[c]{@{}c@{}}No. of \\ Units\end{tabular}} & \textbf{\begin{tabular}[c]{@{}c@{}}No. of Memory \\ Segments\end{tabular}} & \textbf{\begin{tabular}[c]{@{}c@{}}Size of \\ Each Segment\end{tabular}} \\ \hline
\multirow{3}{*}{\textbf{Bus}} & \textbf{Train} & 0.8 & 0.5 & 128 & 10 & 60 \\
 & \textbf{Light Rail} & 0.8 & 0.5 & 128 & 5 & 60 \\
 & \textbf{Ferry} & 0.4 & 0.5 & 128 & 5 & 60 \\ \hline
\multirow{3}{*}{\textbf{Train}} & \textbf{Bus} & 0.8 & 0.5 & 128 & 5 & 60 \\
 & \textbf{Light Rail} & 0.2 & 0.5 & 128 & 5 & 60 \\
 & \textbf{Ferry} & 0.2 & 0.5 & 128 & 5 & 60 \\ \hline
\multirow{3}{*}{\textbf{Light Rail}} & \textbf{Bus} & 0.4 & 0.5 & 128 & 5 & 60 \\
 & \textbf{Train} & 0.8 & 0.5 & 128 & 5 & 60 \\
 & \textbf{Ferry} & 0.2 & 0.5 & 128 & 5 & 60 \\ \hline
\multirow{3}{*}{\textbf{Ferry}} & \textbf{Bus} & 0.6 & 0.5 & 128 & 10 & 60 \\
 & \textbf{Train} & 0.8 & 0.5 & 128 & 5 & 60 \\
 & \textbf{Light Rail} & 0.2 & 0.5 & 64 & 5 & 60 \\ \hline
\end{tabular}}
\label{table: paramter2}
\end{table}

\begin{itemize}
\item\textbf{Historical Average (HA)}: The predicted demand is computed as the average values of historical demand at the same time interval of every day.
\item \textbf{Linear Regression (LR)}: The relations between variables are modeled to minimize the sum of the squares of the errors for forecasting.
\item \textbf{eXtreme Gradient Boosting (Xgboost)} \citep{chen2016xgboost}: Based on the gradient boosting tree, XGBoost is proposed to incorporate the advantages of Bagging integrated learning methods in the evolution process. And the tree-booster is used for demand forecasting.
\item \textbf{Multilayer Perceptron (MLP)}: The neural network contains two fully connected layers to predict the demand.
\item \textbf{Long-Short Term Memory (LSTM)}: LSTM is a variant of Recurrent Neural Network, which is applied to process and analyze time-series information and further predict demand for each transit mode. 
\item \textbf{Dual-stage Attention-based Recurrent Neural Network (DA-RNN)} \citep{qin2017dual}: In the construction of DA-RNN, the input attention mechanism is utilized to extract relevant driving series at each time step while the temporal attention mechanism is utilized to select relevant encoder hidden states across all time steps for time-series prediction. Since the original structure is used for one variant forecasting, we modify it to multiple variants.
\item \textbf{Convolutional LSTM (ConvLSTM)} \citep{xingjian2015convolutional}: ConvLSTM extends the structure of original LSTM to have convolutional structures to capture spatial and temporal correlations simultaneously.
\item \textbf{BiLSTM-ED} \citep{fan2019multi}: This network is proposed based on the framework of Transformer \citep{vaswani2017attention} which consists of the self-attention mechanism and fully connected layers without using recurrent networks or convolution operations.
\item \textbf{Dual Self-Attention Network (DSANet)} \citep{huang2019dsanet}: DSANet is composed of a global temporal convolution module and a local temporal convolution module to capture global and local temporal patterns, respectively. Moreover, the self-attention mechanism is used to model dependencies among various series. The traditional autoregressive linear model is integrated to improve the robustness of the model.
\item \textbf{MT-LSTM}: Two LSTM layers are used to extract temporal correlations for each transport mode independently, and fully connected layers are adopted to analyze the implicit relations for multimodal demand forecasting.
\item \textbf{Knowledge Adaptation with Attentive Multi-task Memory Network (KA2M2)} \citep{li2021multi}: A memory-augmented recurrent network is proposed to capture and store the temporal information of each transit mode. And an attention-based knowledge adaptation module is integrated to adapt relevant knowledge from the station-intensive mode to the station-sparse mode for further forecasting. Note that KA2M2 requires direct data sharing among different transport modes.
\item \textbf{Fine-Tune} \citep{hinton2006fast}: Fine-tune is one of the transfer learning methods for deep learning, which needs a relatively small amount of data. By modifying the structure and selectively loading the weight of the pre-trained network optimized by the source dataset, this method retrains the model with the target dataset.
\end{itemize}

The tested model hyper-parameters are tuned on the validation dataset to locate the best settings for the evaluated deep-based methods for comparison. The hyper-parameter settings of compared models with single-mode (MLP, LSTM, DA-RNN, ConvLSTM, BiLSTM-ED, and DSANet) are summarized in Table~\ref{table: paramter1} and the setting of AK2M2 is summarized in Table~\ref{table: paramter2}.

The performances of all the methods as mentioned above are summarized in Table~\ref{table: overall comparison} and Table~\ref{table:overall2}. For the models dealing with multiple modes (i.e., MT-LSTM, AK2M2, Fine-Tune, and the proposed \textbf{Un-Kadf} framework), the listed results in Table~\ref{table: overall comparison} are the average values of the forecasting results for each mode. Table~\ref{table:overall2} presents the detailed performance of models dealing with various combinations of transport modes. According to these results, we have the following observations.

\begin{table}[h]
\small
\caption{Overall Comparison between the Proposed Method and Existing Methods}
\setlength{\tabcolsep}{.3mm}{
\begin{tabular}{c|ccc|ccc|ccc|ccc}
\hline
\textbf{Mode} & \multicolumn{3}{c|}{\textbf{Bus}} & \multicolumn{3}{c|}{\textbf{Train}} & \multicolumn{3}{c|}{\textbf{Light Rail}} & \multicolumn{3}{c}{\textbf{Ferry}} \\ \hline 
\textbf{Model} & MAE & RMSE & MAPE & MAE & RMSE & MAPE & MAE & RMSE & MAPE & MAE & RMSE & MAPE \\ \hline
\textbf{HA} & 12.231 & 34.489 & 0.359 & 55.424 & 91.315 & 0.350 & 14.927 & 39.703 & 0.347 & 19.681 & 55.739 & 0.407 \\
\textbf{LR} & 11.712 & 23.554 & 0.310 & 53.020 & 85.141 & 0.319 & 13.534 & 31.122 & 0.281 & 19.161 & 49.026 & 0.383 \\
\textbf{Xgboost} & 10.992 & 21.475 & 0.234 & 25.081 & 69.571 & 0.171 & 12.916 & 25.871 & 0.205 & 14.178 & 37.583 & 0.250 \\
\textbf{MLP} & 10.058 & 24.747 & 0.186 & 24.401 & 69.244 & 0.154 & 14.413 & 29.738 & 0.213 & 19.727 & 52.626 & 0.287 \\
\textbf{LSTM} & 8.750 & 20.107 & 0.168 & 24.356 & 70.262 & 0.157 & 12.106 & 25.819 & 0.179 & 14.298 & 37.356 & 0.210 \\
\textbf{DA-RNN} & 8.998 & 20.521 & 0.160 & 23.368 & 65.698 & 0.149 & 11.842 & 22.817 & 0.175 & 13.931 & 36.164 & 0.201 \\
\textbf{ConvLSTM} & 8.391 & 19.081 & 0.149 & 21.263 & 52.529 & 0.143 & 11.844 & 22.887 & 0.168 & 14.040 & 35.171 & 0.181 \\
\textbf{BiLSTM-ED} & 8.051 & 18.964 & 0.151 & 19.903 & 58.102 & 0.127 & 10.885 & 23.531 & 0.165 & 13.113 & 36.800 & 0.192 \\
\textbf{DSANet} & 8.010 & 18.123 & 0.150 & 20.786 & 60.560 & 0.155 & 11.522 & 23.505 & 0.165 & 13.520 & 36.578 & 0.205 \\ \hline
\textbf{MT-LSTM} & 8.540 & 20.344 & 0.165 & 23.074 & 65.624 & 0.147 & 10.448 & 21.280 & 0.162 & 13.495 & 34.266 & 0.201 \\ 
\textbf{AK2M2} & 8.025 & 18.595 & 0.158 & 21.052 & 59.329 & 0.134 & \textbf{9.924} & \textbf{20.669} & \textbf{0.156} & \textbf{12.419} & \textbf{33.485} & \textbf{0.180} \\
\textbf{Fine-Tune} & 8.703 & 21.424 & 0.167 & 23.924 & 66.231 & 0.156 & 11.891 & 23.508 & 0.179 & 15.554 & 39.719 & 0.236 \\ \hline
\textbf{Our Model} & \underline{7.841} & \underline{17.814} & \underline{0.150} & \underline{19.614} & \underline{54.693} & \underline{0.130} & \underline{10.530} & \underline{21.381} & \underline{0.160} & \underline{13.039} & \underline{35.832} & \underline{0.187} \\ \hline
\end{tabular}}
\label{table: overall comparison}
\end{table}

\begin{table}[h!]
\small
\caption{Overall Comparison of Multimodal Transport Forecasting (Single Mode: the forecasting performance obtained by LSTM for single-mode, PCT: growth percentage compared with forecasting performance for single-mode)}
\setlength{\tabcolsep}{.5mm}{
\begin{tabular}{c|c|c|c|cc|cc|cc|cc}
\hline
\textbf{\begin{tabular}[c]{@{}c@{}}Evaluation \\ Matrices\end{tabular}} & \textbf{Mode} & \textbf{\begin{tabular}[c]{@{}c@{}}Single \\ Mode\end{tabular}} & \textbf{\begin{tabular}[c]{@{}c@{}} Cooperative \\ Mode\end{tabular}} & \multicolumn{2}{c|}{\textbf{Fine-Tune}} & \multicolumn{2}{c|}{\textbf{MT-LSTM}} & \multicolumn{2}{c|}{\textbf{Ak2M2}} & \multicolumn{2}{c}{\textbf{My Model}} \\ \hline
 &  &  &  & \textbf{Value} & \textbf{PCT} & \textbf{Value} & \textbf{PCT} & \textbf{Value} & \textbf{PCT} & \textbf{Value} & \textbf{PCT} \\ \hline
\multirow{12}{*}{\textbf{MAE}} & \multirow{3}{*}{\textbf{Bus}} & \multirow{3}{*}{8.750} & \textbf{Train} & 8.689 & 0.70\% & 8.520 & 2.64\% & \textbf{7.714} & 11.85\% & \underline{7.777} & 11.13\% \\
 &  &  & \textbf{Light Rail} & 8.640 & 1.26\% & 8.388 & 4.14\% & 8.096 & 7.48\% & \underline{7.871} & 10.05\% \\
 &  &  & \textbf{Ferry} & 8.780 & -0.34\% & 8.711 & 0.44\% & 8.265 & 5.54\% & \underline{7.874} & 10.01\% \\ \cline{2-12} 
 & \multirow{3}{*}{\textbf{Train}} & \multirow{3}{*}{24.357} & \textbf{Bus} & 24.101 & 1.05\% & 22.745 & 6.62\% & \textbf{18.296} & 24.88\% & \underline{19.564} & 19.68\% \\
 &  &  & \textbf{Light Rail} & 23.183 & 4.82\% & 23.753 & 2.48\% & 22.453 & 7.82\% & \underline{19.724} & 19.02\% \\
 &  &  & \textbf{Ferry} & 24.489 & -0.54\% & 22.725 & 6.70\% & 22.408 & 8.00\% & \underline{19.552} & 19.73\% \\ \cline{2-12} 
 & \multirow{3}{*}{\textbf{\begin{tabular}[c]{@{}c@{}} Light \\ Rail\end{tabular}}} & \multirow{3}{*}{12.106} & \textbf{Bus} & 11.713 & 3.24\% & 10.928 & 9.73\% & \textbf{9.606} & 20.65\% & \underline{10.539} & 12.94\% \\
 &  &  & \textbf{Train} & 11.979 & 1.04\% & 10.361 & 14.41\% & \textbf{9.614} & 22.14\% & \underline{10.468} & 13.52\% \\
 &  &  & \textbf{Ferry} & 11.979 & 1.04\% & 10.055 & 12.84\% & 11.552 & 4.58\% & \underline{10.582} & 12.59\% \\ \cline{2-12} 
 & \multirow{3}{*}{\textbf{Ferry}} & \multirow{3}{*}{14.298} & \textbf{Bus} & 15.715 & -9.91\% & 13.247 & 7.35\% & \textbf{11.917} & 16.65\% & \underline{13.100} & 8.38\% \\
 &  &  & \textbf{Train} & 15.478 & -8.25\% & 13.626 & 4.70\% & \textbf{12.295} & 14.01\% & \underline{12.836} & 10.22\% \\
 &  &  & \textbf{Light Rail} & 15.469 & -8.19\% & 13.613 & 4.79\% & \textbf{13.044} & 8.77\% & \underline{13.179} & 7.82\% \\ \hline
\multirow{12}{*}{\textbf{RMSE}} & \multirow{3}{*}{\textbf{Bus}} & \multirow{3}{*}{20.108} & \textbf{Train} & 22.648 & -12.63\% & 20.805 & -3.47\% & \textbf{17.437} & 13.28\% & \underline{18.409} & 8.46\% \\
 &  &  & \textbf{Light Rail} & 21.173 & -5.30\% & 19.947 & 0.80\% & 19.084 & 5.09\% & \underline{17.355} & 13.69\% \\
 &  &  & \textbf{Ferry} & 20.451 & -1.70\% & 20.281 & -0.86\% & 19.265 & 4.19\% & \underline{17.681} & 10.01\% \\ \cline{2-12} 
 & \multirow{3}{*}{\textbf{Train}} & \multirow{3}{*}{70.262} & \textbf{Bus} & 67.237 & 4.31\% & 64.148 & 8.70\% & \textbf{50.100} & 28.69\% & \underline{55.614} & 20.85\% \\
 &  &  & \textbf{Light Rail} & 65.489 & 6.80\% & 65.230 & 7.16\% & 63.941 & 9.00\% & \underline{54.469} & 22.48\% \\
 &  &  & \textbf{Ferry} & 65.970 & 6.11\% & 67.495 & 3.94\% & 63.946 & 8.99\% & \underline{53.996} & 23.15\% \\ \cline{2-12} 
 & \multirow{3}{*}{\textbf{\begin{tabular}[c]{@{}c@{}} Light \\ Rail\end{tabular}}}& \multirow{3}{*}{25.819} & \textbf{Bus} & 22.995 & 10.94\% & 22.138 & 14.26\% & \textbf{19.851} & 23.12\% & \underline{20.602} & 20.20\% \\
 &  &  & \textbf{Train} & 23.929 & 7.32\% & 20.101 & 22.15\% & \textbf{20.102} & 22.14\% & \underline{21.380} & 17.19\% \\
 &  &  & \textbf{Ferry} & 23.601 & 8.59\% & 21.601 & 16.34\% & \textbf{22.054} & 14.58\% & \underline{22.160} & 14.17\% \\ \cline{2-12} 
 & \multirow{3}{*}{\textbf{Ferry}} & \multirow{3}{*}{37.356} & \textbf{Bus} & 38.545 & -3.18\% & 34.008 & 8.96\% & \textbf{33.880} & 9.30\% & \underline{34.746} & 6.99\% \\
 &  &  & \textbf{Train} & 40.545 & -8.54\% & 34.803 & 6.83\% & \textbf{34.113} & 8.68\% & \underline{35.432} & 5.15\% \\
 &  &  & \textbf{Light Rail} & 40.068 & -7.26\% & 34.186 & 8.49\% & \textbf{32.460} & 13.10\% & \underline{37.318} & 0.10\% \\ \hline
\multirow{12}{*}{\textbf{MAPE}} & \multirow{3}{*}{\textbf{Bus}} & \multirow{3}{*}{0.169} & \textbf{Train} & 0.166 & 1.78\% & 0.163 & 3.55\% & \textbf{0.145} & 12.42\% & \underline{0.148} & 12.43\% \\
 &  &  & \textbf{Light Rail} & 0.166 & 1.78\% & 0.163 & 3.55\% & 0.161 & 2.96\% & \underline{0.151} & 10.65\% \\
 &  &  & \textbf{Ferry} & 0.169 & 0.00\% & 0.169 & 0.00\% & 0.168 & 0.59\% & \underline{0.151} & 10.65\% \\ \cline{2-12} 
 & \multirow{3}{*}{\textbf{Train}} & \multirow{3}{*}{0.157} & \textbf{Bus} & 0.154 & 1.91\% & 0.144 & 8.28\% & \textbf{0.124} & 21.02\% & \underline{0.129} & 17.84\% \\
 &  &  & \textbf{Light Rail} & 0.150 & 4.46\% & 0.146 & 7.00\% & 0.141 & 10.19\% & \underline{0.131} & 16.56\% \\
 &  &  & \textbf{Ferry} & 0.164 & -4.46\% & 0.152 & 3.18\% & 0.138 & 12.10\% & \underline{0.130} & 17.20\% \\ \cline{2-12} 
 & \multirow{3}{*}{\textbf{\begin{tabular}[c]{@{}c@{}} Light \\ Rail\end{tabular}}} & \multirow{3}{*}{0.180} & \textbf{Bus} & 0.179 & 0.56\% & 0.159 & 11.67\% & \textbf{0.154} & 14.44\% & \underline{0.161} & 10.56\% \\
 &  &  & \textbf{Train} & 0.177 & 1.67\% & 0.154 & 14.44\% & \textbf{0.150} & 16.67\% & \underline{0.158} & 12.22\% \\
 &  &  & \textbf{Ferry} & 0.179 & 0.56\% & 0.172 & 4.44\% & 0.164 & 8.89\% & \underline{0.162} & 10.00\% \\ \cline{2-12} 
 & \multirow{3}{*}{\textbf{Ferry}} & \multirow{3}{*}{0.210} & \textbf{Bus} & 0.248 & -18.10\% & 0.196 & 6.67\% & \textbf{0.169} & 19.52\% & \underline{0.188} & 10.48\% \\
 &  &  & \textbf{Train} & 0.230 & -9.52\% & 0.203 & 3.33\% & \textbf{0.175} & 16.67\% & \underline{0.182} & 13.33\% \\
 &  &  & \textbf{Light Rail} & 0.231 & -10.00\% & 0.205 & 2.38\% & 0.197 & 6.19\% & \underline{0.191} & 9.05\% \\ \hline
\end{tabular}}
\label{table:overall2}
\end{table}

First of all, classical machine learning methods HA and LR obtain higher values in terms of MAE, RMSE, and MAPE than other strategies. The forecasting performance of Xgboost is better than HA, LR, and some deep learning models in some sources (e.g., more accurate than MLP and LSTM for the ferry), which implies the effectiveness of ensemble methods.

Second, the deep-learning-based strategies normally perform better than HA and LR since neural networks can handle non-linear relations of large-scale demand data more thoroughly. And among the tested deep models, MLP gains less satisfactory results due to its limited ability for temporal information extraction. When compared to MLP, LSTM achieves higher accuracy on four modes, demonstrating the usefulness of the recurrent network, which is used as the fundamental component for demand forecasting in the current study.

Furthermore, DA-RNN, as a state-of-the-art strategy, does not obtain better results than LSTM on the demand forecasting of the bus with abundant stations since it is proposed for uni-variant time-series prediction. It realizes a competent performance on the other three modes (i.e., train, light rail, and ferry) with fewer stations. In addition, RNN-based and CNN-based models (i.e., ConvLSTM and DSANet) proposed for demand forecasting with the data of one transport mode yield relatively accurate results, implying their ability to handle temporal patterns via recurrent networks and spatial patterns via convolutional networks. BiLSTM-ED, a Transformer-based neural network, shows its superior capability to distinguish efficacious travel patterns from the raw demand data for further prediction. It is of our interest to further explore methods based on Transformer for multimodal demand forecasting in the future.

The strategies designed based on graph convolution networks (GCNs) have not been tested and compared in the experiments since the station-based demand data is relatively sparse for graph construction. Training on large-scale sparse networks by GCN remains challenging \citep{yadav2019lovasz}, which is not the research focus of this work. However, GCN-based models have become powerful tools for transport analysis in numerous studies, which will be our future direction to study spatial-temporal correlations for unsupervised knowledge adaptation and transport prediction.

Moreover, MT-LSTM optimizes two modes simultaneously for demand forecasting, which performs better than LSTM training with one mode and demonstrates that the correlations/similarities between two modes can boost prediction tasks. AK2M2 is proposed for multi-mode transport forecasting, adapting the learned knowledge from the station-intensive mode to the station-sparse mode via optimizing two datasets simultaneously. AK2M2 achieves more comprehensive performance than the proposed model \textbf{Un-Kadf} on the average results of the light rail and the ferry, but works worse on the bus and the train as shown in Table~\ref{table: overall comparison}. AK2M2 is proposed for station-sparse mode (light rail and ferry) forecasting improvement and requires simultaneous access to all data sources, it is expected that it should perform better regarding station-sparse mode. However, while the proposed method in this study does not require simultaneous access to all data sources, it produces very competitive forecasting results for station-sparse mode (light rail and ferry) and performs better (on average) than AK2M2 for the other two modes. Also, our work gains more accurate results on some pairs of modes involving station-sparse mode (e.g., light rail as the target dataset and the ferry as the source dataset) as listed in Table~\ref{table:overall2}. These results verify the potential of the proposed unsupervised knowledge adaptation that avoids access to detailed source data and the requirement of data sharing.

Fine-tune, as a popular and basic transfer learning strategy, achieves better performance on the bus, train, and light rail than LSTM but works worse than other state-of-the-art models. It indicates that utilizing the pre-trained model optimized by the source dataset is able to improve the forecasting performance of the target dataset.

In general, compared to the listed state-of-the-art methods training with single-mode forecasting, MT-LSTM, and fine-tune, \textbf{Un-Kadf} yields improvement in all the three evaluation matrices on four transit modes. When compared to AK2M2, it performs better on the bus and train, and worse but very close on the light rail and ferry. However, AK2M2 requires direct utilization of multiple datasets simultaneously for training to some extent, while the proposed \textbf{Un-Kadf} avoids direct data sharing. 

Overall, our results indicate that using the information recorded in the pre-trained network introduced in Subsection~\ref{sec:pretrain} optimized by the source dataset can efficiently enhance the forecasting performance of the target dataset via the model sharing network discussed in Subsection~\ref{sec:sharing} based on the strategy of unsupervised knowledge adaptation instead of depending on the data sharing.

\begin{comment}
In addition, to evaluate the computation cost, we summarize the training time of one epoch for the proposed \textbf{Un-Kadf} and other seven deep models (MLP, LSTM, DA-RNN, ConvLST, BiLSTM-ED, DSANet, and AK2M2) on the bus dataset in Table~\ref{table:time}. As can be seen, though the proposed method runs slightly slower than MLP, LSTM, and BiLSTM-ED, it achieves better forecasting performance than them. Also, our model takes less time on training than DA-RNN, ConvLSTM, DSANet, and AK2M2. Therefore, considering the significant performance improvement as shown in Table~\ref{table: overall comparison}, the computation cost of \textbf{Un-Kadf} is moderate.
\begin{table}[]
\small
\caption{Comparison of Computation Efficiency on the Bus Dataset}
\setlength{\tabcolsep}{.4mm}{
\begin{tabular}{c|c|c|c|c|c|c|c|c}
\hline
\textbf{Model} & \textbf{MLP} & \textbf{LSTM} & \textbf{DA-RNN} & \textbf{ConvLSTM} & \textbf{BiLSTM-ED} & \textbf{DSANet} & \textbf{AK2M2} & \textbf{Un-Kadf} \\ \hline
\textbf{\begin{tabular}[c]{@{}c@{}}Training \\ Time (epoch)\end{tabular}} & 0.057 & 0.069 & 0.978 & 0.556 & 0.132 & 3.746 & 2.980 & 0.158 \\ \hline
\end{tabular}}
\label{table:time}
\end{table}
\end{comment}

\section{Conclusion} \label{sec:conclusion}

This study demonstrates that unsupervised knowledge/information adaptation from the source transport mode to the target transport mode via model sharing is able to improve passenger demand prediction, where the Unsupervised Knowledge Adaptation Enhanced Demand Forecasting (\textbf{Un-Kadf}) framework has been proposed. In particular, we propose to extract the source knowledge based on an encoder-decoder framework and record the knowledge in the pre-trained recurrent network. Then, the pre-trained network is adapted to the target dataset instead of direct data sharing to enhance the forecasting performance since similarities/relations among multiple transport modes are helpful for demand prediction. The evaluation results on real-world datasets collected from Sydney, including four public transit modes (bus, train, light rail, and ferry), show that the proposed approach is able to improve the forecasting performance of the target dataset and achieve more accurate results.

Although the proposed methodology is tested on a public multimodal transport system, it can be adapted to systems with other modes such as private cars and ride-hailing vehicles. It can also be used for the same transport mode managed by different institutions or operated in different cities. The zone-based and city-based demand prediction problems can also be accommodated by the proposed model in this study by replacing the station-based demand setting with zone/region-based demand setting. Generally, this study provides insights on using model sharing instead of the dependence on direct data sharing to improve demand prediction performance in multimodal transport systems.

In the future, this work can be extended in several lines. For instance, motivated by the success of CNN-based and GCN-based models for demand prediction, the spatial-temporal relations among various stations or regions can be studied and further adapted via unsupervised knowledge adaptation strategies. Also, other types of features, such as weather and Point of Interest information, can be investigated for forecasting enhancement via moderate modifications of the proposed model.

%\section*{Acknowledgments}

\appendix

\section{Mechanism of LSTM} \label{sec:LSTM}

Owing to the scheme of the memory unit and gate mechanism \citep{hochreiter1997long}, long-term dependencies can be extracted and exploding/vanishing problems can be solved compared to classical Recurrent Neural Network (RNN) by LSTM, motivating the heavily usage in natural language process area and further in transport prediction problems. In detail, LSTM consists of an input gate $\mathbf{i}^{t} \in \mathbb{R}^{m}$, a forget gate $\mathbf{f}^{t} \in \mathbb{R}^{m}$, an output gate $\mathbf{o}^{t} \in \mathbb{R}^{m}$, an internal memory cell $\mathbf{c}^{t} \in \mathbb{R}^{m}$, and a hidden state $\mathbf{h}^{t} \in \mathbb{R}^{m}$ where $m$ is the dimension of the hidden state. The derivation of LSTM layer is represented as:
\begin{equation}
\begin{aligned}
    & \mathbf{i}^{t} = \sigma (\mathbf{W}_{i} \mathbf{x}^{t} + \mathbf{U}_{i} \mathbf{h}^{t-1} + \mathbf{b}_{i}) \\
    & \mathbf{f}^{t} = \sigma (\mathbf{W}_{f} \mathbf{x}^{t} + \mathbf{U}_{f} \mathbf{h}^{t-1} + \mathbf{b}_{f}) \\
    & \mathbf{o}^{t} = \sigma (\mathbf{W}_{o} \mathbf{x}^{t} + \mathbf{U}_{o} \mathbf{h}^{t-1} + \mathbf{b}_{o}) \\
    & \mathbf{\theta} ^{t} = \tanh (\mathbf{W}_{\theta} \mathbf{x}^{t} + \mathbf{U}_{\theta} \mathbf{h}^{t-1} + \mathbf{b}_{\theta}) \\
    & \mathbf{c}^{t} = \mathbf{f}^{t} \otimes \mathbf{c}^{t-1} + \mathbf{i}^{t} \otimes \mathbf{\theta} ^{t} \\
    & \mathbf{h}^{t} = \mathbf{o}^{t} \otimes \tanh (\mathbf{c}^{t})
\end{aligned}
\label{formula:lstm}
\end{equation}
where $\otimes$ represents element-wise multiplication, $\sigma$ denotes the logistic sigmoid function $\sigma(u) = 1 / (1 + e^{-u})$, $\mathbf{x}^{t} \in \mathbb{R}^{n}$ is the input vector at the current time step $t$ ($n$ denotes the input dimension), and $\mathbf{W}_{i},\mathbf{W}_{f},\mathbf{W}_{o},\mathbf{W}_{\theta} \in \mathbb{R}^{m \times n}$ and $\mathbf{U}_{i},\mathbf{U}_{f},\mathbf{U}_{o},\mathbf{U}_{\theta} \in \mathbb{R}^{m \times m}$ are weight matrices, and $\mathbf{b}_{i},\mathbf{b}_{f},\mathbf{b}_{o},\mathbf{b}_{\theta} \in \mathbb{R}^{m}$ are bias vectors. And we abbreviate the input data and output data of the LSTM layer as $\mathbf{h}^{t}, \mathbf{c}^{t} =  LSTM (\mathbf{x}^{t}, \mathbf{h}^{t-1}, \mathbf{c}^{t-1})$.

\bibliographystyle{apalike}

\bibliography{cas-refs}

\begin{thebibliography}{}

\bibitem[Bai et~al., 2019]{bai2019stg2seq}
Bai, L., Yao, L., Kanhere, S.~S., Wang, X., and Sheng, Q.~Z. (2019).
\newblock {STG2seq}: spatial-temporal graph to sequence model for multi-step
  passenger demand forecasting.
\newblock In {\em 28th International Joint Conference on Artificial
  Intelligence (IJCAI 2019)}, pages 1981--1987.

\bibitem[Bai et~al., 2020]{bai2020adaptive}
Bai, L., Yao, L., Li, C., Wang, X., and Wang, C. (2020).
\newblock Adaptive graph convolutional recurrent network for traffic
  forecasting.
\newblock {\em Advances in Neural Information Processing Systems},
  33:17804--17815.

\bibitem[Chen and Guestrin, 2016]{chen2016xgboost}
Chen, T. and Guestrin, C. (2016).
\newblock Xgboost: A scalable tree boosting system.
\newblock In {\em Proceedings of the 22nd ACM SIGKDD International Conference
  on Knowledge Discovery and Data Mining}, pages 785--794.

\bibitem[Chen et~al., 2020]{chen2020nonconvex}
Chen, X., Yang, J., and Sun, L. (2020).
\newblock A nonconvex low-rank tensor completion model for spatiotemporal
  traffic data imputation.
\newblock {\em Transportation Research Part C: Emerging Technologies},
  117:102673.

\bibitem[Cheng et~al., 2022]{cheng2022real}
Cheng, Z., Trepanier, M., and Sun, L. (2022).
\newblock Real-time forecasting of metro origin-destination matrices with
  high-order weighted dynamic mode decomposition.
\newblock {\em Transportation Science}, pages 1--15.

\bibitem[Cui et~al., 2020]{cui2020learning}
Cui, Z., Ke, R., Pu, Z., Ma, X., and Wang, Y. (2020).
\newblock Learning traffic as a graph: A gated graph wavelet recurrent neural
  network for network-scale traffic prediction.
\newblock {\em Transportation Research Part C: Emerging Technologies},
  115:102620.

\bibitem[Devlin et~al., 2019]{devlin2019bert}
Devlin, J., Chang, M., Lee, K., and Toutanova, K. (2019).
\newblock Bert: Pre-training of deep bidirectional transformers for language
  understanding.
\newblock In {\em Proceedings of the 2019 Conference of the North American
  Chapter of the Association for Computational Linguistics: Human Language
  Technologies}, pages 4171--4186.

\bibitem[Fan et~al., 2019]{fan2019multi}
Fan, C., Zhang, Y., Pan, Y., Li, X., Zhang, C., Yuan, R., Wu, D., Wang, W.,
  Pei, J., and Huang, H. (2019).
\newblock Multi-horizon time series forecasting with temporal attention
  learning.
\newblock In {\em Proceedings of the 25th ACM SIGKDD International Conference
  on Knowledge Discovery \& Data Mining}, pages 2527--2535.

\bibitem[Feng et~al., 2018]{feng2018adaptive}
Feng, X., Ling, X., Zheng, H., Chen, Z., and Xu, Y. (2018).
\newblock Adaptive multi-kernel svm with spatial--temporal correlation for
  short-term traffic flow prediction.
\newblock {\em IEEE Transactions on Intelligent Transportation Systems},
  20(6):2001--2013.

\bibitem[Finn and Levine, 2018]{finn2018meta}
Finn, C. and Levine, S. (2018).
\newblock Meta-learning and universality: Deep representations and gradient
  descent can approximate any learning algorithm.
\newblock In {\em International Conference on Learning Representations}.

\bibitem[Geng et~al., 2019]{geng2019spatiotemporal}
Geng, X., Li, Y., Wang, L., Zhang, L., Yang, Q., Ye, J., and Liu, Y. (2019).
\newblock Spatiotemporal multi-graph convolution network for ride-hailing
  demand forecasting.
\newblock In {\em Proceedings of the AAAI Conference on Artificial
  Intelligence}, volume~33, pages 3656--3663.

\bibitem[Guo et~al., 2020]{guo2020optimized}
Guo, K., Hu, Y., Qian, Z., Liu, H., Zhang, K., Sun, Y., Gao, J., and Yin, B.
  (2020).
\newblock Optimized graph convolution recurrent neural network for traffic
  prediction.
\newblock {\em IEEE Transactions on Intelligent Transportation Systems},
  22(2):1138--1149.

\bibitem[Guo et~al., 2019]{guo2019deep}
Guo, S., Lin, Y., Li, S., Chen, Z., and Wan, H. (2019).
\newblock Deep spatial--temporal 3d convolutional neural networks for traffic
  data forecasting.
\newblock {\em IEEE Transactions on Intelligent Transportation Systems},
  20(10):3913--3926.

\bibitem[Hinton et~al., 2006]{hinton2006fast}
Hinton, G.~E., Osindero, S., and Teh, Y.~W. (2006).
\newblock A fast learning algorithm for deep belief nets.
\newblock {\em Neural Computation}, 18(7):1527--1554.

\bibitem[Hochreiter and Schmidhuber, 1997]{hochreiter1997long}
Hochreiter, S. and Schmidhuber, J. (1997).
\newblock Long short-term memory.
\newblock {\em Neural Computation}, 9(8):1735--1780.

\bibitem[Huang et~al., 2019]{huang2019dsanet}
Huang, S., Wang, D., Wu, X., and Tang, A. (2019).
\newblock Dsanet: Dual self-attention network for multivariate time series
  forecasting.
\newblock In {\em Proceedings of the 28th ACM International Conference on
  Information and Knowledge Management}, pages 2129--2132.

\bibitem[Jin et~al., 2020]{jin2020urban}
Jin, G., Cui, Y., Zeng, L., Tang, H., Feng, Y., and Huang, J. (2020).
\newblock Urban ride-hailing demand prediction with multiple spatio-temporal
  information fusion network.
\newblock {\em Transportation Research Part C: Emerging Technologies},
  117:102665.

\bibitem[Ke et~al., 2021]{ke2021joint}
Ke, J., Feng, S., Zhu, Z., Yang, H., and Ye, J. (2021).
\newblock Joint predictions of multi-modal ride-hailing demands: A deep
  multi-task multi-graph learning-based approach.
\newblock {\em Transportation Research Part C: Emerging Technologies},
  127:103063.

\bibitem[Ke et~al., 2018]{ke2018hexagon}
Ke, J., Yang, H., Zheng, H., Chen, X., Jia, Y., Gong, P., and Ye, J. (2018).
\newblock Hexagon-based convolutional neural network for supply-demand
  forecasting of ride-sourcing services.
\newblock {\em IEEE Transactions on Intelligent Transportation Systems},
  20(11):4160--4173.

\bibitem[Kim et~al., 2020]{kim2020stepwise}
Kim, T., Sharda, S., Zhou, X., and Pendyala, R.~M. (2020).
\newblock A stepwise interpretable machine learning framework using linear
  regression (lr) and long short-term memory (lstm): City-wide demand-side
  prediction of yellow taxi and for-hire vehicle (fhv) service.
\newblock {\em Transportation Research Part C: Emerging Technologies},
  120:102786.

\bibitem[Li et~al., 2020a]{li2020graph}
Li, C., Bai, L., Liu, W., Yao, L., and Waller, S.~T. (2020a).
\newblock Graph neural network for robust public transit demand prediction.
\newblock {\em IEEE Transactions on Intelligent Transportation Systems}, pages
  1--13.

\bibitem[Li et~al., 2020b]{li2020knowledge}
Li, C., Bai, L., Liu, W., Yao, L., and Waller, S.~T. (2020b).
\newblock Knowledge adaption for demand prediction based on multi-task memory
  neural network.
\newblock In {\em Proceedings of the 29th ACM International Conference on
  Information \& Knowledge Management}, pages 715--724.

\bibitem[Li et~al., 2021a]{li2021multi}
Li, C., Bai, L., Liu, W., Yao, L., and Waller, S.~T. (2021a).
\newblock A multi-task memory network with knowledge adaptation for multimodal
  demand forecasting.
\newblock {\em Transportation Research Part C: Emerging Technologies},
  131:103352.

\bibitem[Li et~al., 2021b]{li2021transferability}
Li, J., Guo, F., Sivakumar, A., Dong, Y., and Krishnan, R. (2021b).
\newblock Transferability improvement in short-term traffic prediction using
  stacked lstm network.
\newblock {\em Transportation Research Part C: Emerging Technologies},
  124:102977.

\bibitem[Li et~al., 2021c]{li2021domain}
Li, J., Zhang, K., Shen, L., Wang, Z., Guo, F., Angeloudis, P., Chen, X.~M.,
  and Hu, S. (2021c).
\newblock A domain adaptation framework for short-term traffic prediction.
\newblock In {\em 2021 IEEE International Intelligent Transportation Systems
  Conference (ITSC)}, pages 3564--3569. IEEE.

\bibitem[Li et~al., 2019]{li2019learning}
Li, Y., Zhu, Z., Kong, D., Xu, M., and Zhao, Y. (2019).
\newblock Learning heterogeneous spatial-temporal representation for
  bike-sharing demand prediction.
\newblock In {\em Proceedings of the AAAI Conference on Artificial
  Intelligence}, volume~33, pages 1004--1011.

\bibitem[Lippi et~al., 2013]{lippi2013short}
Lippi, M., Bertini, M., and Frasconi, P. (2013).
\newblock Short-term traffic flow forecasting: An experimental comparison of
  time-series analysis and supervised learning.
\newblock {\em IEEE Transactions on Intelligent Transportation Systems},
  14(2):871--882.

\bibitem[Liu et~al., 2019]{liu2019contextualized}
Liu, L., Qiu, Z., Li, G., Wang, Q., Ouyang, W., and Lin, L. (2019).
\newblock Contextualized spatial--temporal network for taxi origin-destination
  demand prediction.
\newblock {\em IEEE Transactions on Intelligent Transportation Systems},
  20(10):3875--3887.

\bibitem[Lv et~al., 2014]{lv2014traffic}
Lv, Y., Duan, Y., Kang, W., Li, Z., and Wang, F. (2014).
\newblock Traffic flow prediction with big data: a deep learning approach.
\newblock {\em IEEE Transactions on Intelligent Transportation Systems},
  16(2):865--873.

\bibitem[Ma and Qian, 2018]{ma2018estimating}
Ma, W. and Qian, Z.~S. (2018).
\newblock Estimating multi-year 24/7 origin-destination demand using
  high-granular multi-source traffic data.
\newblock {\em Transportation Research Part C: Emerging Technologies},
  96:96--121.

\bibitem[Ma et~al., 2018]{ma2018parallel}
Ma, X., Zhang, J., Du, B., Ding, C., and Sun, L. (2018).
\newblock Parallel architecture of convolutional bi-directional lstm neural
  networks for network-wide metro ridership prediction.
\newblock {\em IEEE Transactions on Intelligent Transportation Systems},
  20(6):2278--2288.

\bibitem[Moreira-Matias et~al., 2013]{moreira2013predicting}
Moreira-Matias, L., Gama, J., Ferreira, M., Mendes-Moreira, J., and Damas, L.
  (2013).
\newblock Predicting taxi--passenger demand using streaming data.
\newblock {\em IEEE Transactions on Intelligent Transportation Systems},
  14(3):1393--1402.

\bibitem[Pan and Yang, 2009]{pan2009survey}
Pan, S.~J. and Yang, Q. (2009).
\newblock A survey on transfer learning.
\newblock {\em IEEE Transactions on Knowledge and Data Engineering},
  22(10):1345--1359.

\bibitem[Peng et~al., 2021]{peng2021cnn}
Peng, Y., Liang, T., Hao, X., Chen, Y., Li, S., and Yi, Y. (2021).
\newblock Cnn-gru-am for shared bicycles demand forecasting.
\newblock {\em Computational Intelligence and Neuroscience}, 2021.

\bibitem[Qin et~al., 2017]{qin2017dual}
Qin, Y., Song, D., Cheng, H., Cheng, W., Jiang, G., and Cottrell, G.~W. (2017).
\newblock A dual-stage attention-based recurrent neural network for time series
  prediction.
\newblock In {\em Proceedings of the 26th International Joint Conference on
  Artificial Intelligence}, pages 2627--2633. AAAI Press.

\bibitem[Tan and Le, 2019]{tan2019efficientnet}
Tan, M. and Le, Q. (2019).
\newblock Efficientnet: Rethinking model scaling for convolutional neural
  networks.
\newblock In {\em International Conference on Machine Learning}, pages
  6105--6114.

\bibitem[Tang et~al., 2021]{tang2021multi}
Tang, J., Liang, J., Liu, F., Hao, J., and Wang, Y. (2021).
\newblock Multi-community passenger demand prediction at region level based on
  spatio-temporal graph convolutional network.
\newblock {\em Transportation Research Part C: Emerging Technologies},
  124:102951.

\bibitem[Toman et~al., 2020]{toman2020dynamic}
Toman, P., Zhang, J., Ravishanker, N., and Konduri, K.~C. (2020).
\newblock Dynamic predictive models for ridesourcing services in new york city
  using daily compositional data.
\newblock {\em Transportation Research Part C: Emerging Technologies},
  121:102833.

\bibitem[Vaswani et~al., 2017]{vaswani2017attention}
Vaswani, A., Shazeer, N., Parmar, N., Uszkoreit, J., Jones, L., Gomez, A.~N.,
  Kaiser, {\L}., and Polosukhin, I. (2017).
\newblock Attention is all you need.
\newblock In {\em Advances in Neural Information Processing Systems}, pages
  5998--6008.

\bibitem[Wang et~al., 2019]{wang2019cross}
Wang, L., Geng, X., Ma, X., Liu, F., and Yang, Q. (2019).
\newblock Cross-city transfer learning for deep spatio-temporal prediction.
\newblock In {\em Proceedings of the 28th International Joint Conference on
  Artificial Intelligence}, pages 1893--1899.

\bibitem[Xingjian et~al., 2015]{xingjian2015convolutional}
Xingjian, S., Chen, Z., Wang, H., Yeung, D.~Y., Wong, W.~K., and Woo, W.~C.
  (2015).
\newblock Convolutional lstm network: A machine learning approach for
  precipitation nowcasting.
\newblock In {\em Advances in Neural Information Processing Systems}, pages
  802--810.

\bibitem[Xu et~al., 2017]{xu2017real}
Xu, J., Rahmatizadeh, R., B{\"o}l{\"o}ni, L., and Turgut, D. (2017).
\newblock Real-time prediction of taxi demand using recurrent neural networks.
\newblock {\em IEEE Transactions on Intelligent Transportation Systems},
  19(8):2572--2581.

\bibitem[Xue et~al., 2015]{xue2015short}
Xue, R., Sun, D.~J., and Chen, S. (2015).
\newblock Short-term bus passenger demand prediction based on time series model
  and interactive multiple model approach.
\newblock {\em Discrete Dynamics in Nature and Society}, 2015.

\bibitem[Yadav et~al., 2019]{yadav2019lovasz}
Yadav, P., Nimishakavi, M., Yadati, N., Vashishth, S., Rajkumar, A., and
  Talukdar, P. (2019).
\newblock Lovasz convolutional networks.
\newblock In {\em The 22nd International Conference on Artificial Intelligence
  and Statistics}, pages 1978--1987.

\bibitem[Yao et~al., 2019]{yao2019learning}
Yao, H., Liu, Y., Wei, Y., Tang, X., and Li, Z. (2019).
\newblock Learning from multiple cities: A meta-learning approach for
  spatial-temporal prediction.
\newblock In {\em The World Wide Web Conference}, pages 2181--2191.

\bibitem[Yao et~al., 2020]{yao2020unsupervised}
Yao, Z., Wang, Y., Long, M., and Wang, J. (2020).
\newblock Unsupervised transfer learning for spatiotemporal predictive
  networks.
\newblock In {\em International Conference on Machine Learning}, pages
  10778--10788.

\bibitem[Ye et~al., 2019]{ye2019co}
Ye, J., Sun, L., Du, B., Fu, Y., Tong, X., and Xiong, H. (2019).
\newblock Co-prediction of multiple transportation demands based on deep
  spatio-temporal neural network.
\newblock In {\em Proceedings of the 25th ACM SIGKDD International Conference
  on Knowledge Discovery \& Data Mining}, pages 305--313.

\bibitem[Zhang et~al., 2021]{zhang2021short}
Zhang, J., Che, H., Chen, F., Ma, W., and He, Z. (2021).
\newblock Short-term origin-destination demand prediction in urban rail transit
  systems: A channel-wise attentive split-convolutional neural network method.
\newblock {\em Transportation Research Part C: Emerging Technologies},
  124:102928.

\bibitem[Zhang et~al., 2019]{zhang2019multistep}
Zhang, Z., Li, M., Lin, X., Wang, Y., and He, F. (2019).
\newblock Multistep speed prediction on traffic networks: A deep learning
  approach considering spatio-temporal dependencies.
\newblock {\em Transportation Research Part C: Emerging Technologies},
  105:297--322.

\bibitem[Zhou et~al., 2021]{zhou2021urban}
Zhou, F., Li, L., Zhang, K., and Trajcevski, G. (2021).
\newblock Urban flow prediction with spatial--temporal neural odes.
\newblock {\em Transportation Research Part C: Emerging Technologies},
  124:102912.

\end{thebibliography}
\end{document}